\documentclass[10pt,twocolumn,letterpaper]{article}

\usepackage{cvpr}
\usepackage{times}
\usepackage{epsfig}
\usepackage[caption=false]{subfig}
\usepackage{graphicx}
\usepackage{amsmath}
\usepackage{amssymb}

\usepackage{soul}
\usepackage{adjustbox}

\usepackage[breaklinks=true,bookmarks=false]{hyperref}
\usepackage[capitalise,noabbrev]{cleveref}
\crefname{equation}{Eq.}{Eqs.}%

\usepackage{textcomp}
\usepackage{gensymb}
\usepackage{comment}
\usepackage{multirow}
\usepackage{chngcntr}

\DeclareMathOperator{\softmax}{softmax}
\DeclareMathOperator{\diag}{diag}
\DeclareMathOperator{\prob}{P}
\newcommand*\boldell{\boldsymbol{\ell}}

\cvprfinalcopy 

\def\cvprPaperID{6066} 

\begin{document}

\definecolor{revcolor}{rgb}{1.0,0.0,0.0}
\newcommand{\rev}[1]{{\textcolor{revcolor}{#1}}}

\newcommand{\acceptedrev}[1]{#1}

\definecolor{sarahcolor}{rgb}{0.858,0.188,0.478}
\newcommand{\sarah}[1]{{\textcolor{sarahcolor}{[SP #1]}}}
\newcommand\sap[1] {\sarah{#1}}

\definecolor{gregcolor}{rgb}{0.758,0.488,0.178}
\newcommand{\greg}[1]{{\textcolor{gregcolor}{[GS #1]}}}
\newcommand\gre[1] {\greg{#1}}

\definecolor{stevencolor}{rgb}{0.1,0.1,0.9}
\newcommand{\steven}[1]{{\textcolor{stevencolor}{[SGM #1]}}}
\newcommand\sgm[1] {\steven{#1}}

\definecolor{danielcolor}{rgb}{0.1,0.9,0.1}
\newcommand{\daniel}[1]{{\textcolor{danielcolor}{[DH #1]}}}
\newcommand\dah[1] {\daniel{#1}}

\definecolor{alescolor}{rgb}{0.4,0.1,0.8}
\newcommand{\ales}[1]{{\textcolor{alescolor}{[AL #1]}}}
\newcommand\al[1] {\ales{#1}}

\definecolor{bencolor}{rgb}{0.0,0.58431372549019607843137254901961,0.65098039215686274509803921568627}
\newcommand{\ben}[1]{\textcolor{bencolor}{[Ben:~{#1}]}}

\definecolor{eduardocolor}{rgb}{0.0,0.4,0.7}
\newcommand{\edu}[1]{{\textcolor{eduardocolor}{[EP #1]}}}
\newcommand\ep[1] {\edu{#1}}

\newcommand{\norm}[1]{\lVert #1 \rVert}

\title{A Multi-Hypothesis Approach to Color Constancy}


\author{Daniel Hernandez-Juarez$^{1}$, \,\,
Sarah Parisot$^{1,2}$, \,\,
Benjamin Busam$^{1,3}$, \,\,
Ale\v{s} Leonardis$^{1}$ \\ 
Gregory Slabaugh$^{1}$ \,\,
Steven McDonagh$^{1}$ \\
{\tt\small dhernandez0@gmail.com,}\\
{\tt\small \{sarah.parisot, benjamin.busam, ales.leonardis, gregory.slabaugh, steven.mcdonagh\}} 
{\tt\small @huawei.com}
\and
$^{1}$Huawei Noah's Ark Lab\\
\and
$^{2}$Mila, Montr\'{e}al
\and
$^{3}$Technical University of Munich
} 

\maketitle

\begin{abstract}





Contemporary approaches frame the color constancy problem as learning camera specific illuminant mappings. While high accuracy can be achieved on camera specific data, these models depend on camera spectral sensitivity and typically exhibit poor generalisation to new devices. Additionally, regression methods produce point estimates that do not explicitly account for potential ambiguities among plausible illuminant solutions, due to the ill-posed nature of the problem.
We propose a Bayesian framework that naturally handles color constancy ambiguity via a multi-hypothesis strategy. Firstly, we select a set of candidate scene illuminants in a data-driven fashion and apply them to a target image to generate of set of corrected images. Secondly, we estimate, for each corrected image, the likelihood of the light source being achromatic using a camera-agnostic CNN. Finally, our method explicitly learns a final illumination estimate from the generated posterior probability distribution. 
Our likelihood estimator learns to answer a camera-agnostic question and thus enables effective multi-camera training by disentangling illuminant estimation from the supervised learning task. We extensively evaluate our proposed approach and additionally set a benchmark for novel sensor generalisation without re-training. Our method provides state-of-the-art accuracy on multiple public datasets (\acceptedrev{up to $11$\% median angular error improvement}) while maintaining real-time execution.

\end{abstract}

\section{Introduction}
\label{sec:intro}

\begin{figure}[th]
\centering
\includegraphics[width=1\columnwidth]{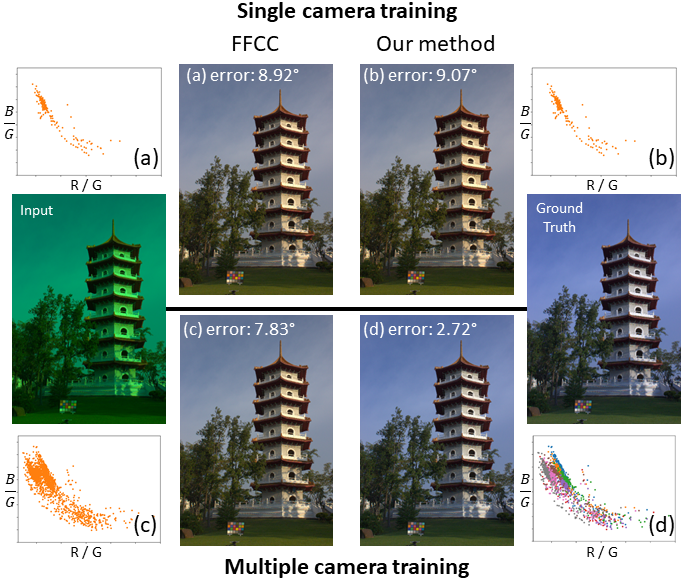}
\caption{Our multi-hypothesis strategy allows us to leverage multi-camera datasets. Example image \acceptedrev{taken} from the NUS dataset~\cite{cheng2014illuminant}. Single camera training: (a) state of the art method FFCC~\cite{barron2017fast} and (b) our method obtains similar angular-error. Training with all $8$ dataset cameras: aggregate all images to (c) define FFCC histogram center and (d) use an illuminant candidate set per camera. [$\frac{r}{g}$, $\frac{b}{g}$] color space plots show training set illuminant distributions. Each camera is encoded with a different color in (d) to highlight camera-specific illuminants. Our model leverages the extra data to achieve lower angular error. \acceptedrev{Images are rendered in sRGB color space}.}
\label{fig:teaser}
\end{figure}




Color constancy is an essential part of digital image processing pipelines. When treated as a computational process, this involves estimation of scene light source color, present at capture time, and correcting an image such that its appearance matches that of the scene captured under an achromatic light source. The algorithmic process of recovering the illuminant of a scene is commonly known as computational Color Constancy (CC) or Automatic White Balance (AWB). Accurate estimation is essential for visual aesthetics~\cite{gijsenij2009perceptual}, as well as downstream high-level computer vision tasks~\cite{Afifi2019WBEmulation,AndreopoulosT12,abs-1803-07721,DiamondSBWH17} that typically require color-unbiased and device-independent images.

Under the prevalent assumption that the scene is illuminated by a single or dominant light source,  the observed pixels of an image are typically modelled using the physical model of Lambertian image formation captured under a trichromatic photosensor:

 \begin{equation}
    \rho_k({X}) = \int_{\Omega} E(\lambda) S(\lambda, X) C_k(\lambda) d \lambda \quad k \in \{R, G, B\}.
    \label{eq:per_pixel_appearance}
\end{equation}

\noindent where $\rho_k(X)$ is the intensity of color channel $k$ at pixel location $X$, $\lambda$ the wavelength of light such that $E(\lambda)$ represents the spectrum of the illuminant, $S(\lambda, X)$ the surface reflectance at pixel location $X$ and $C_k(\lambda)$ camera sensitivity function for channel $k$, considered over the spectrum of wavelengths $\Omega$.


The goal of computational CC then becomes estimation of the global illumination color $\rho^E_k$ where:

\begin{equation}
    \rho^E_k = \int_{\Omega} E(\lambda) C_k(\lambda) d \lambda \hspace{5mm} k \in \{R, G, B\}.
    \label{eq:illuminant_eq}
\end{equation}
Finding $\rho^E_k$ in~\cref{eq:illuminant_eq} results in a ill-posed problem due to the existence of infinitely many combinations of illuminant and surface reflectance that result in identical observations at each pixel $X$.


A natural and popular solution for learning-based color constancy is to frame the problem as a regression task \cite{afifi2019bmvc,hu2017fc4,gong2019bmvc,bianco2017single,shi2016deep,lou2015color,bianco2015color}. However, typical regression methods provide a point estimate and do not offer any information regarding possible alternative solutions. Solution ambiguity is present in many vision domains~\cite{rupprecht2017learning,manhardt2019ambiguity} and is particularly problematic in the cases where multi-modal solutions exist~\cite{MahendranAV18}. 
Specifically for color constancy we note that, due to the ill-posed nature of the problem, multiple illuminant solutions are often possible with varying probability. Data-driven approaches that learn to directly estimate the illuminant result in learning tasks that are inherently camera-specific due to the camera sensitivity function \cf \cref{eq:illuminant_eq}. This observation will often manifest as a sensor domain gap; models trained on a single device typically exhibit poor generalisation to novel cameras.







In this work, we propose to address the ambiguous nature of the color constancy problem through multiple hypothesis estimation. Using a Bayesian formulation, we discretise the illuminant space and estimate the likelihood that each considered illuminant accurately corrects the observed image. We evaluate how plausible an image is after illuminant correction, and gather a discrete set of plausible solutions in the illuminant space. This strategy can be interpreted as framing color constancy as a classification problem, similar to recent promising work in this direction~\cite{barron2015convolutional,barron2017fast,OhK17}. Discretisation strategies have also been successfully employed in other computer vision domains, such as 3D pose estimation \cite{MahendranAV18} and object detection~\cite{RedmonDGF16,RenHG017}, resulting in \eg state of the art accuracy improvement. 



In more detail, we propose to decompose the AWB task into three sub-problems: a) selection of a set of \emph{candidate} illuminants b) learning to estimate the likelihood that an image, corrected by a candidate, is illuminated achromatically, and c) combining candidate illuminants, using the estimated posterior probability distribution, to produce a final output.


We correct an image with all candidates independently and evaluate the likelihood of each solution with a shallow CNN. 
Our network learns to estimate the likelihood of \emph{white balance correctness} for a given image. In contrast to prior work, we disentangle camera-specific illuminant estimation from the learning task thus allowing to train a single, device agnostic, AWB model that can effectively leverage multi-device data. We avoid distribution shift and resulting domain gap problems~\cite{afifi2019bmvc,nguyen2014raw,gao2017improving}, associated with camera specific training, and propose a well-founded strategy to leverage multiple data. Principled combination of datasets is of high value for learning based color constancy given the typically small nature of individual color constancy datasets (on the order of only hundreds of images). See~\cref{fig:teaser}.
\newline
\noindent Our \textbf{contributions} can be summarised as: 
\begin{enumerate}
    \item We decompose the AWB problem into a novel \textbf{multi-hypothesis three stage} pipeline. 
    
    
    
    \item We introduce a \textbf{multi-camera learning} strategy that allows to leverage multi-device datasets and improve accuracy over single-camera training.
    
    
    \item We provide a \textbf{training-free} model adaptation strategy for new cameras. 
    
    \item We report improved \textbf{state-of-the-art performance} on two popular public datasets (NUS~\cite{cheng2014illuminant}, Cube+~\cite{BanicL18a}) and competitive results on Gehler-Shi~\cite{shi2000re,gehler2008bayesian}.
    
\end{enumerate}

\begin{figure*}[th]
\begin{center}
   \includegraphics[width=1\linewidth]{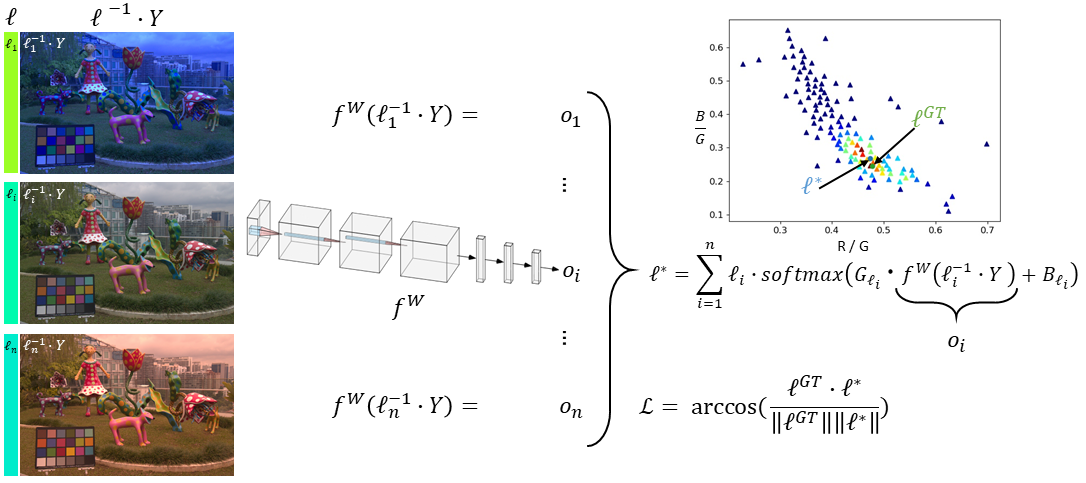}
\end{center}
   \caption{Method overview: we first generate a list of $n$ candidate illuminants $\boldell_i$ (candidate illuminants are shown left of the respective corrected images) using $K$-means clustering~\cite{Lloyd82}. We correct the input image with each of the $n$ candidates independently and then estimate the likelihood $o_i$ of each corrected image with our network. We combine illuminant candidates using the posterior probability distribution to generate an illuminant estimation $\boldell^*$. The error is back-propagated through the network using angular error loss $\mathcal{L}$. The [$\frac{r}{g}$, $\frac{b}{g}$] plot in the upper-right illustrates the posterior probability distribution (triangles encoded from blue to red) of the candidates $\boldell_i$, the final prediction vector $\boldell^*$ (blue circle) and the ground-truth illuminant $\boldell^{GT}$ (green circle). \acceptedrev{Images are rendered in sRGB color space}.}
\label{fig:method}
\end{figure*}


\section{Related work}
\label{sec:related_work}

\textbf{Classical color constancy} methods utilise low-level statistics to realise various instances of the gray-world assumption: \emph{the average reflectance in a scene under a neutral light source is achromatic}. Gray-World~\cite{buchsbaum1980spatial} and its extensions \cite{Finlayson2004,van2007edge} are based on these assumptions that tie scene reflectance statistics (\eg mean, max reflectance) to the achromaticity of scene color. 

Related assumptions define perfect reflectance~\cite{land1971lightness,funt2010rehabilitation} and result in White-Patch methods. Statistical methods are fast and typically contain few free parameters, however their performance is highly dependent on strong scene content assumptions and these methods falter in cases where these assumptions fail to hold.

An early \textbf{Bayesian framework}~\cite{FreemanB95} used Bayes' rule to compute the posterior distribution for the illuminants and scene surfaces. They model the prior of the illuminant and the surface reflectance as a truncated multivariate normal distribution on the weights of a linear model. 
Other Bayesian works~\cite{RosenbergML03,gehler2008bayesian}, discretise the illuminant space and model the surface reflectance priors by learning real world histogram frequencies; in \cite{RosenbergML03} the prior is modelled as a uniform distribution over a subset of illuminants while \cite{gehler2008bayesian} uses the empirical distribution of the training illuminants. Our work uses the Bayesian formulation proposed in previous works~\cite{RosenbergML03,FreemanB95,gehler2008bayesian}. We estimate the likelihood probability distribution with a CNN which also explicitly learns to model the prior distribution for each illuminant.


\textbf{Fully supervised methods.}
Early learning-based works~\cite{funt2004estimating,xiong2006estimating,wang2009edge} comprise combinational and direct approaches, typically relying on hand-crafted image features which limited their overall performance. Recent fully supervised convolutional color constancy work offers state-of-the-art estimation accuracy. Both local patch-based~\cite{bianco2015color,shi2016deep,bianco2017single} and full image input~\cite{barron2015convolutional,lou2015color,barron2017fast,gong2019bmvc,hu2017fc4} have been considered, investigating different model architectures~\cite{bianco2015color,bianco2017single,shi2016deep} and the use of semantic information~\cite{hu2017fc4,lou2015color,barron2017fast}. 

Some methods frame color constancy as a \textbf{classification problem}, \eg CCC \cite{barron2015convolutional} and the follow-up refinement FFCC~\cite{barron2017fast}, by using a color space that identifies image re-illumination with a histogram shift. Thus, they elegantly and efficiently evaluate different illuminant candidates. Our method also discretises the illuminant space but we explicitly select the candidate illuminants, allowing for multi-camera training while FFCC~\cite{barron2017fast} is constrained to use all histogram bins as candidates and single-camera training.

\acceptedrev{The method of~\cite{OhK17} uses $K$-means~\cite{Lloyd82} to cluster illuminants of the dataset and then applies a CNN to frame the problem as a classification task; network input is a single (pre-white balanced) image and output results in $K$ class probabilities, representing the prospect of each illuminant (each class) explaining the correct image illumination. 
Our method first chooses candidate illuminants similarly, however, the key difference is that our model learns to infer whether an image is \emph{well white balanced or not}. We ask this question $K$ times by correcting the image, independently, with each illuminant candidate. 
This affords an independent estimation of the likelihood for each illuminant and thus enables multi-device training to improve results.}

\textbf{Multi-device training}
The method of~\cite{afifi2019bmvc} introduces a two CNN approach; the first network learns a `sensor independent' linear transformation ($3{\times}3$ matrix), the RGB image is transformed to this `canonical' color space and then, a second network provides the predicted illuminant. The method is trained on multiple datasets except the test camera and obtains competitive results.


The work of~\cite{McDonagh2018} affords fast adaptation to previously unseen cameras, and robustness to changes in capture device by leveraging annotated samples across different cameras and datasets in a meta-learning framework.


A recent approach~\cite{bianco2019quasi}, makes an assumption that sRGB images collected from the web are well white balanced, therefore, they apply a simple \acceptedrev{de-gamma correction to approximate an inverse tone mapping} and then find achromatic pixels with a CNN to predict the illuminant. These web images were captured with unknown cameras, were processed by different ISP pipelines and might have been modified with image editing software. Despite additional assumptions, the method achieves promising results, however, not comparable with the supervised state-of-the-art.


\acceptedrev{In contrast we propose an alternative technique to enable multi-camera training and mitigate well understood sensor domain-gaps. We can train a single CNN using images captured by different cameras through the use of \emph{camera-dependent} illuminant candidates. This property, of accounting for camera-dependent illuminants, affords fast model adaption; accurate inference is achievable for images captured by cameras not seen during training, if camera illuminant candidates are available (removing the need for model re-training or fine-tuning). We provide further methodological detail of these contributions and evidence towards their efficacy in Sections~\ref{sec:method} and~\ref{sec:results} respectively.}


\section{Method}
\label{sec:method}

Let $\mathbf{y} = (y_r,y_g,y_b)$ be a pixel from an input image $Y$ in linear RGB space. We model the global illumination, \cref{eq:illuminant_eq}, with the standard linear model~\cite{von1905influence} such that each pixel $\mathbf{y}$ is the product of the surface reflectance $\mathbf{r} = (r_r,r_g,r_b)$ and a global illuminant $\boldell = (\ell_r,\ell_g,\ell_b)$ shared by all pixels such that:

\begin{equation}
y_k = r_k \cdot \ell_k \hspace{5mm} k \in \{R, G, B\}.
\label{eq:image_formation}
\end{equation}
Given $Y= (\mathbf{y}_1,\ldots,\mathbf{y}_m)$, comprising $m$ pixels, and $R = (\mathbf{r}_1,\ldots,\mathbf{r}_m)$, our goal is to estimate $\boldell$ and produce $R = \diag(\boldell)^{-1} Y$.

In order to estimate the correct illuminant to adjust the input image $Y$, we propose to frame the CC problem with a probabilistic generative model with unknown surface reflectances and illuminant. 
We consider a set $\boldell_i \in \mathbb{R}^3, i \in \{1,\ldots,n\}$ of candidate illuminants, each of which are applied to $Y$ to generate a set of $n$ tentatively corrected images $\diag(\boldell_i)^{-1} Y$. Using the set of corrected images as inputs, we then train a CNN to identify the most probable illuminants such that the final estimated illuminant is a linear combination of the candidates. In this section, we first introduce our general Bayesian framework, followed by our proposed implementation of the main building blocks of the model. 
An overview of the method can be seen in \cref{fig:method}. 

\subsection{Bayesian approach to color constancy}


Following the Bayesian formulation previously considered~\cite{RosenbergML03,FreemanB95,gehler2008bayesian}, we assume that the color of the light and the surface reflectance are independent. Formally $\prob(\boldell,R){=}\prob(\boldell) \prob(R)$, \ie knowledge of the surface reflectance provides us with no additional information about the illuminant, $\prob(\, \boldell \mid R){=}\prob(\boldell)$. Based on this assumption we decompose these factors and model them separately. 




Using Bayes' rule, we define the posterior distribution of $\boldell$ illuminants given the input image $Y$ as:

\begin{equation}
    \prob(\,\boldell \mid Y) = \frac {\prob(\, Y \mid \boldell \,) \prob(\boldell)}{\prob(Y)}.
    \label{eq:posterior}
\end{equation}

We model the likelihood of an observed image $Y$ for a given illuminant $\boldell$:

\begin{equation}
\begin{split}
    \prob(\, Y \mid \boldell \,) = \int_r \prob(\, Y \mid \boldell, R = r) \prob(R = r) \,dr \\
            = \prob(R = \diag(\boldell)^{-1} Y)
    \end{split}
    \label{eq:likelihood}
\end{equation}
where $R$ are the surface reflectances and $\diag(\boldell)^{-1} Y$ is the image as corrected with illuminant $\boldell$. The term $\prob(\, Y \mid \boldell, R = r)$ is only non-zero for $R=\diag(\boldell)^{-1} Y$. The likelihood rates whether a corrected image looks realistic. 

We choose to instantiate the model of our likelihood using a shallow CNN. The network should learn to output a high likelihood if the reflectances look realistic. We model the prior probability $\prob(\boldell)$ for each candidate illuminant independently as learnable parameters in an end-to-end approach; this effectively acts as a regularisation, favouring more likely real-world illuminants. We note that, in practice, the function modelling the prior also depends on factors such as the environment (indoor / outdoor), the time of day, ISO etc. However, the size of currently available datasets prevent us from modelling more complex proxies.



In order to estimate the illuminant $\boldell^*$, we optimise the quadratic cost (minimum MSE Bayesian estimator), minimised by the mean of the posterior distribution:

\begin{equation}
    \boldell^* = \int_{\boldell} \boldell \cdot \prob(\, \boldell \mid Y) \,d\boldell
\label{eq:mean}
\end{equation}

This is done in the following three steps (\cf \cref{fig:method}):
\begin{enumerate}
    \item \textbf{Candidate selection} (\cref{subsec:candidate_selection}): Choose a set of $n$ illuminant candidates to generate $n$ corrected thumbnail ($64{\times}64$) images. 
    \item \textbf{Likelihood estimation} (\cref{subsec:likelihood_estimation}): Evaluate these $n$ images independently with a CNN, a network designed to estimate the likelihood that an image is well white balanced $\prob(Y \mid \boldell)$. 
    \item \textbf{Illuminant determination} (\cref{subsec:illuminant_determination}): Compute the posterior probability of each candidate illuminant and determine a final illuminant estimation $\boldell^*$. 
\end{enumerate}

This formulation allows estimation of a posterior probability distribution, allowing us to reason about a set of \emph{probable} illuminants rather than produce a single illuminant point estimate (\cf regression approaches). Regression typically does not provide feedback on a possible set of alternative solutions which has shown to be of high value in alternative vision problems~\cite{MahendranAV18}.

The second benefit that our decomposition affords is a principled multi-camera training process. A single, device agnostic CNN estimates illuminant likelihoods and performs independent selection of candidate illuminants for each camera. By leveraging image information across multiple datasets we increase model robustness. Additionally, the amalgamation of small available CC datasets provides a step towards harnessing the power of large capacity models for this problem domain \cf contemporary models.

\subsection{Candidate selection}
\label{subsec:candidate_selection}





The goal of candidate selection is to discretise the illuminant space of a specific camera in order to obtain a set of representative illuminants (spanning the illuminant space). Given a collection of ground truth illuminants, measured from images containing calibration objects (\ie a labelled training set), we compute candidates using $K$-means clustering~\cite{Lloyd82} on the linear RGB space.

By forming $n$ clusters of our measured illuminants, we define the set of candidates $\boldell_i \in \mathbb{R}^3, i \in \{1,\ldots,n\}$ as the cluster centers. $K$-means illuminant clustering is previously shown to be effective for color constancy~\cite{OhK17} however we additionally evaluate alternative candidate selection strategies (detailed in the supplementary material); our experimental investigation confirms a simple $K$-means approach provides strong target task performance. Further, the effect of $K$ is empirically evaluated \acceptedrev{in~\cref{subsec:qualitative_evaluation}}. 

Image $Y,$ captured by a given camera, is then used to produce a set of images, corrected using the illuminant candidate set for the camera, on which we evaluate the accuracy of each candidate.

\subsection{Likelihood estimation}
\label{subsec:likelihood_estimation}

We model the likelihood estimation step using a neural network which, for a given illuminant $\boldell$ and image $Y$, takes the tentatively corrected image $\diag(\boldell)^{-1} Y$ as input, and learns to predict the likelihood $P(Y \vert \boldell)$ that the image has been well white balanced \ie has an appearance of being captured under an achromatic light source.



The success of low capacity histogram based methods~\cite{barron2015convolutional,barron2017fast} and the inference-training tradeoff for small datasets motivate a compact network design. We propose a small CNN with one spatial convolution and subsequent layers constituting $1{\times}1$ convolutions with spatial pooling. Lastly, three fully connected layers gradually reduce the dimensionality to one (see supplementary material for architecture details). Our network output is then a single value that represents the log-likelihood that the image is well white balanced:
\begin{equation}
    \log(\prob(\, Y \mid \boldell \,)) = f^{W}(\diag(\boldell)^{-1} Y \,).
    \label{eq:cnn}
\end{equation}





Function \acceptedrev{$f^{W}$} is our trained CNN parametrised by model weights $W$. \cref{eq:cnn} estimates the log-likelihood of each candidate illuminant separately. 
\acceptedrev{It is important to note that we only train a single CNN which is used to estimate the likelihood for each candidate illuminant independently.} However, in practice, certain candidate illuminants will be more common than others. 
To account for this, following~\cite{barron2017fast}, we compute an affine transformation of our log-likelihood $\log(\prob(\, Y \mid \boldell \,))$ by introducing learnable, illuminant specific, gain $G_{\boldell}$ and bias $B_{\boldell}$ parameters. Gain $G_l$ affords amplification of illuminant likelihoods. 
The bias term $B_{\boldell}$ learns to prefer some illuminants \ie a prior distribution in a Bayesian sense: $B_{\boldell} = \log(\prob(\boldell))$. The log-posterior probability can then be formulated as:

\begin{equation}
  \log(\prob(\, \boldell \mid Y \,)) = G_{\boldell} \cdot \log(\prob(\, Y \mid \boldell \,)) + B_{\boldell}.
  \label{eq:affine_transformation}
\end{equation}

\acceptedrev{We highlight that learned affine transformation parameters are training camera-dependent and provide further discussion on camera agnostic considerations in~\cref{subsec:multidevice_training}}.


\subsection{Illuminant determination}
\label{subsec:illuminant_determination}

We require a differentiable method in order to train our model end-to-end, and therefore the use of a simple Maximum a Posteriori (MAP) inference strategy is not possible. Therefore to estimate the illuminant $\boldell^*$, we use the minimum mean square error Bayesian estimator, which is minimised by the posterior mean of $\boldell$ (\cf \cref{eq:mean}): 




\begin{equation}
\begin{split}
    \boldell^* = \sum_{i=1}^{n} \boldell_i \cdot \softmax(\log(\prob(\boldell_i \mid Y))) \\
    = \frac{1}{ \sum e^{\log(\prob(\boldell_i \mid Y))} } \sum_{i=1}^{n} \boldell_i \cdot e^{\log(\prob(\boldell_i \mid Y))} .
    \end{split}
    \label{eq:final_prediction}
\end{equation}


\acceptedrev{The resulting vector $\boldell^*$ is $l_2$-normalised.} We leverage our $K$-means centroid representation of the linear RGB space and use linear interpolation within the convex hull of feasible illuminants to determine the estimated scene illuminant $\boldell^*$. 
For \cref{eq:final_prediction}, we take inspiration from~\cite{KendallMDH17,OhK17}, who have successfully explored similar strategies \acceptedrev{in CC and stereo regression}, \eg \cite{KendallMDH17} introduced an analogous \emph{soft-argmin} to estimate disparity values from a set of candidates. We apply a similar strategy for illuminant estimation and use the \emph{soft-argmax} which provides a linear combination of all candidates weighted by their probabilities.




We train our network end-to-end with the commonly used angular error loss function, where $\boldell^*$ and $\boldell^{GT}$ are the prediction and ground truth illuminant, respectively:

\begin{equation}
    \mathcal{L}_{error} = \arccos(\frac{\boldell^{GT} \cdot \boldell^*}{\norm{\boldell^{GT}} \norm{\boldell^*}})
    \label{eq:angular_error}
\end{equation}

\subsection{Multi-device training}
\label{subsec:multidevice_training}

\acceptedrev{As discussed in previous work~\cite{afifi2019bmvc,nguyen2014raw,gao2017improving},} CC models typically fail to train successfully using multiple camera data due to distribution shifts between camera sensors, making them intrinsically device-dependent and limiting model capacity. A device-independent model is highly appealing due to the small number of images commonly available in camera-specific public color constancy datasets. The cost and time associated with collecting and labelling new large data for specific novel devices is expensive and prohibitive.

Our CNN learns to produce the likelihood that an input image is well white balanced. We claim that framing part of the CC problem in this fashion results in a device-independent learning task. We evaluate the benefit of this hypothesis experimentally in~\cref{sec:results}.

To train with multiple cameras we use camera-specific candidates, yet learn only a single model. Specifically, we train with a different camera for each batch, use camera-specific candidates yet update a single set of CNN parameters during model training. \acceptedrev{In order to ensure that our CNN is device-independent, we fix previously learnable parameters that depend on sensor specific illuminants, \ie $B_{\boldell} = 0$ and $G_{\boldell} = 1$.} 
The absence of these parameters, learned in a camera-dependent fashion, intuitively restricts model flexibility however we observe this drawback to be compensated by the resulting ability to train using amalgamated multi-camera datasets \ie more data. This strategy allows our CNN to be camera-agnostic and affords the option to refine existing CNN quality when data from novel cameras becomes available. We however clarify that our overarching strategy for white balancing maintains use of camera-specific candidate illuminants. 


\section{Results}
\label{sec:results}

\subsection{Training details}
\label{subsec:training_details}

We train our models for $120$ epochs and use $K$-mean~\cite{Lloyd82} with $K{=}120$ candidates. Our batch size is $32$, we use the Adam optimiser~\cite{kingma2014adam} with initial learning rate $5{\times}10^{-3}$, divided by two after $10$, $50$ and $80$ epochs. Dropout~\cite{dropout2012} of 50\% is applied after average pooling. \acceptedrev{We take the log transform of the input before the first convolution.} Efficient inference is feasible by concatenating each candidate corrected image into the batch dimension. We use PyTorch 1.0~\cite{steiner2019pytorch} and an Nvidia Tesla V100 for our experiments.
The first layer is the only spatial convolution, it is adapted from~\cite{SimonyanZ14a} and pretrained on ImageNet \cite{imagenet_cvpr09}. We fix the weights of this first layer to avoid over-fitting. The total amount of weights is $22.8K$.
For all experiments calibration objects are masked, black level subtracted and over-saturated pixels are clipped at $95\%$ threshold. We resize the image to $64{\times}64$ and normalise.

\subsection{Datasets}
\label{subsec:datasets}

We experiment using three public datasets. The \textbf{Gehler-Shi} dataset~\cite{shi2000re,gehler2008bayesian} contains $568$ images of indoor and outdoor scenes. Images were captured using \emph{Canon 1D} and \emph{Canon 5D} cameras. We highlight our awareness of the existence of multiple sets of non-identical ground-truth labels for this dataset (see~\cite{hemrit2018rehabilitating} for further detail). Our Gehler-Shi evaluation is conducted using the SFU ground-truth labels~\cite{shi2000re} (consistent with the label naming convention in~\cite{hemrit2018rehabilitating}). The \textbf{NUS} dataset~\cite{cheng2014illuminant} originally consists of $8$ subsets of $\sim$210 images per camera providing a total of $1736$ images. The \textbf{Cube+} dataset~\cite{BanicL18a} contains $1707$ images captured with \emph{Canon 550D} camera, consisting of predominantly outdoor imagery.


For the NUS~\cite{cheng2014illuminant} and Gehler-Shi~\cite{shi2000re,gehler2008bayesian} datasets we perform three-fold cross validation (CV) \acceptedrev{using the splits provided in previous work~\cite{barron2017fast,barron2015convolutional}}. The Cube+~\cite{BanicL18a} dataset does not provide splits for CV so we use all images for learning and evaluate using a related set of test images, provided for the recent Cube+ ISPA 2019 challenge~\cite{ISPAChallenge}. We compare with the results from the challenge leader-board.

For the NUS dataset~\cite{cheng2014illuminant}, we additionally explore training \emph{multi-camera} models \acceptedrev{and thus create a new set of CV folds to facilitate this}. We are careful to highlight that the NUS dataset consists of eight image subsets, pertaining to eight capture devices. Each of our new folds captures a distinct set of scene content (\ie sets of up to eight similar images for each captured scene). This avoids testing on \emph{similar scene content} seen during training. 
We define our multi-camera CV such that multi-camera fold $i$ is the concatenation of images, pertaining to common scenes, captured from all eight cameras. 
\acceptedrev{The folds that we define are made available in our supplementary material.}

\subsection{Evaluation metrics}

We use the standard angular error metric for quantitative evaluation (\cf \cref{eq:angular_error}). 
We report standard CC statistics to summarise results over the investigated datasets: Mean, Median, Trimean, Best $25\%$, Worst $25\%$. We further report method inference time in the supplementary material. Other works' results were taken from corresponding papers, resulting in missing statistics for some methods.
The NUS~\cite{cheng2014illuminant} dataset is composed of $8$ cameras, we report the geometric mean of each statistic for each method across all cameras as standard in the literature~\cite{barron2017fast,barron2015convolutional,hu2017fc4}.

\subsection{Quantitative evaluation}
\label{subsec:qualitative_evaluation}
\noindent
\textbf{Accuracy experiments.}
We report competitive results on the dataset of Gehler-Shi~\cite{shi2000re,gehler2008bayesian} (\cf \cref{table:gehler_shi_results}). This dataset can be considered very challenging as the number of images per camera is imbalanced: There are $86$ \emph{Canon 1D} and $482$ \emph{Canon 5D} images. Our method is not able to outperform the state-of-the-art likely due to the imbalanced nature and small size of \emph{Canon 1D}. Pretraining on a combination of NUS~\cite{cheng2014illuminant} and Cube+~\cite{BanicL18a} provides moderate accuracy improvement despite the fact that the Gehler-Shi dataset has a significantly different illuminant distribution compared to those seen during pre-training. \acceptedrev{We provide additional experiments, exploring the effect of varying $K$, for $K$-means candidate selection in the supplementary material.}


\begin{table}
\begin{adjustbox}{max width=\columnwidth}
\begin{tabular}{|l|c|c|c|c|c|}
\hline
Method & Mean & Med. & Tri. & Best 25\% & Worst 25\% \\
\hline\hline
Gray-world~\cite{buchsbaum1980spatial}         & 6.36 & 6.28 & 6.28 & 2.33 & 10.58 \\ 
White-Patch~\cite{brainard1986analysis}        & 7.55 & 5.86 & 6.35 & 1.45 & 16.12 \\ 
Bayesian~\cite{gehler2008bayesian}  & 4.82 & 3.46 & 3.88 & 1.26 & 10.49 \\ 
Quasi-unsupervised~\cite{bianco2019quasi} & 2.91 & 1.98 & - & - & - \\
Afifi \etal 2019~\cite{afifi2019bmvc} & 2.77 & 1.93 & - & 0.55 & 6.53 \\
Meta-AWB~\cite{McDonagh2018} & 2.57 & 1.84 & 1.94 & 0.47 &  6.11 \\
Cheng \etal 2015~\cite{cheng2015effective} & 2.42 & 1.65 & 1.75 & 0.38 & 5.87 \\
CM 2019~\cite{gong2019bmvc} & 2.48 & 1.61 & 1.80 & 0.47 & 5.97 \\
Oh \etal~\cite{OhK17}        & 2.16 & 1.47 & 1.61 & 0.37 & 5.12 \\
CCC~\cite{barron2015convolutional} & 1.95 & 1.22 & 1.38 & 0.35 & 4.76 \\
DS-Net~\cite{shi2016deep} & 1.90 & 1.12 & 1.33 & 0.31 & 4.84 \\
FC4~\cite{hu2017fc4} (SqueezeNet) & 1.65 & 1.18 & 1.27 & 0.38 & \textbf{3.78} \\
FC4~\cite{hu2017fc4} (AlexNet) & 1.77 & 1.11 & 1.29 & 0.34 & 4.29 \\
FFCC~\cite{barron2017fast} (model P) & \textbf{1.61} & \textbf{0.86} & \textbf{1.02} & \textbf{0.23} & 4.27 \\
\hline
Ours & 2.35 & 1.43 & 1.63 & 0.40 & 5.80 \\
Ours (pretrained) & 2.10 & 1.32 & 1.53 & 0.36 & 5.10 \\
\hline
\end{tabular}
\end{adjustbox}
\caption{Angular error statistics for Gehler-Shi dataset~\cite{shi2000re,gehler2008bayesian}.}
\label{table:gehler_shi_results}
\end{table}




\acceptedrev{Results for NUS~\cite{cheng2014illuminant} are provided in~\cref{table:nus8_results}. Our method obtains competitive accuracy and the previously observed trend, pre-training using additional datasets (here Gehler-Shi~\cite{shi2000re,gehler2008bayesian} and Cube+~\cite{BanicL18a}), again improves results.}

\acceptedrev{In~\cref{table:nus8_multicam_results}, we report results for our multi-device setting on the NUS~\cite{cheng2014illuminant} dataset. For this experiment we introduce a new set of training folds to ensure that scenes are well separated and refer to Sections~\ref{subsec:multidevice_training} for multi-device training and~\ref{subsec:datasets} for related training folds detail. 
We draw multi-device comparison with FFCC~\cite{barron2017fast}, by choosing to center the FFCC histogram with the training set (of amalgamated camera datasets). Note that results are not directly comparable with~\cref{table:nus8_results} due to our redefinition of CV folds.}
\acceptedrev{
Our method is more accurate than the state-of-the-art when training considers all available cameras at the same time. Note that multi-device training improves the median angular error of each individual camera dataset (we provide results in the supplementary material). Overall performance is improved by ${\sim}11\%$ in terms of median accuracy.} 

\begin{table}
\begin{adjustbox}{max width=\columnwidth}
\begin{tabular}{|l|c|c|c|c|c|}
\hline
Method & Mean & Med. & Tri. & Best 25\% & Worst 25\% \\
\hline\hline
White-patch~\cite{brainard1986analysis} & 9.91 & 7.44 & 8.78 & 1.44 & 21.27 \\
Gray-world~\cite{buchsbaum1980spatial} & 4.59 & 3.46 & 3.81 & 1.16 & 9.85 \\
Bayesian~\cite{gehler2008bayesian}  & 3.50 & 2.36 & 2.57 & 0.78 & 8.02 \\ 
Oh \etal~\cite{OhK17}        & 2.36 & 2.09 & - & - & \textbf{4.16} \\
Quasi-unsupervised~\cite{bianco2019quasi} & 1.97 & 1.91 & - & - & - \\
CM 2019~\cite{gong2019bmvc} & 2.25 & 1.59 & 1.74 & 0.50 & 5.13 \\
FC4~\cite{hu2017fc4} (SqueezeNet) & 2.23 & 1.57 & 1.72 & 0.47 & 5.15 \\
FC4~\cite{hu2017fc4} (AlexNet) & 2.12 & 1.53 & 1.67 & 0.48 & 4.78 \\
Afifi \etal 2019~\cite{afifi2019bmvc} & 2.05 & 1.50 & - & 0.52 & 4.48 \\
CCC~\cite{barron2015convolutional} & 2.38 & 1.48 & 1.69 & 0.45 & 5.85 \\
Cheng \etal 2015~\cite{cheng2015effective} & 2.18 & 1.48 & 1.64 & 0.46 & 5.03 \\
DS-Net~\cite{shi2016deep} & 2.21 & 1.46 & 1.68 & 0.48 & 6.08 \\
Meta-AWB~\cite{McDonagh2018} & \textbf{1.89} & 1.34 & 1.44 & 0.45 & 4.28 \\
FFCC~\cite{barron2017fast} (model Q) & 2.06 & 1.39 & 1.53 & 0.39 & 4.80 \\
FFCC~\cite{barron2017fast} (model M) & 1.99 & \textbf{1.31} & \textbf{1.43} & \textbf{0.35} & 4.75 \\
\hline
Ours & 2.39 & 1.61 & 1.74 & 0.50 & 5.67 \\
Ours (pretrained) & 2.35 & 1.55 & 1.73 & 0.46 & 5.62 \\
\hline
\end{tabular}
\end{adjustbox}
\caption{Angular error statistics for NUS~\cite{cheng2014illuminant}.}
\label{table:nus8_results}
\end{table}

\begin{table}
\begin{adjustbox}{max width=\columnwidth}
\begin{tabular}{|l|c|c|c|c|c|}

\hline
Method & Mean & Med. & Tri. & Best 25\% & Worst 25\% \\
\hline\hline
\multicolumn{6}{|l|}{\textbf{One model per device}} \\
\hline
FFCC~\cite{barron2017fast} (model Q) & 2.37 & 1.50 & 1.69 & 0.46 & 5.76 \\
Ours (pretrained) & 2.35 & 1.48 & 1.67 & 0.47 & 5.71 \\
\hline
\multicolumn{6}{|l|}{\textbf{Multi-device training}} \\
\hline
FFCC~\cite{barron2017fast} (model Q) & 2.59 & 1.77 & 1.94 & 0.52 & 6.14 \\
Ours (pretrained) & 2.22 & 1.33 & 1.53 & 0.44 & 5.49 \\
\hline
\end{tabular}
\end{adjustbox}
\caption{Angular error statistics for NUS~\cite{cheng2014illuminant} using multi-device cross-validation folds (see \cref{subsec:datasets}). FFCC \emph{model Q} is considered for fair comparison (thumbnail resolution input).}
\label{table:nus8_multicam_results}
\end{table}

We also outperform the state-of-the-art on the recent Cube challenge~\cite{ISPAChallenge} as shown in \cref{table:cube_test_results}. Pretraining together on Gehler-Shi~\cite{shi2000re,gehler2008bayesian} and NUS~\cite{cheng2014illuminant} improves our \textit{Mean} and \textit{Worst 95\%} statistics. 

In summary, we observe strong generalisation when using multiple camera training (\eg NUS~\cite{cheng2014illuminant} results \cf \cref{table:nus8_results,table:nus8_multicam_results}). 
These experiments illustrate the large benefit achievable with multi-camera training when illuminant distributions of the cameras are broadly consistent. 
Gehler-Shi~\cite{shi2000re,gehler2008bayesian} has a very disparate illuminant distribution with respect to alternative datasets and we are likely unable to exploit the full advantage of multi-camera training. We note the FFCC~\cite{barron2017fast} state of the art method is extremely shallow and therefore optimised for small datasets. In contrast, when our model is trained on large and relevant datasets we are able to achieve superior results. 
\newline
\newline
\textbf{Run time.}
Regarding run-time; we measure inference speed at ${\sim}10$ milliseconds, implemented in unoptimised PyTorch (see supplementary material for further detail). 
\newline
\begin{table}
\begin{adjustbox}{max width=\columnwidth}
\begin{tabular}{|l|c|c|c|c|c|}
\hline
Method & Mean & Med. & Tri. & Best 25\% & Worst 25\% \\
\hline\hline
Gray-world~\cite{buchsbaum1980spatial} & 4.44 & 3.50 & - & 0.77 & 9.64 \\
1st-order Gray-Edge~\cite{van2007edge} & 3.51 & 2.30 & - & 0.56 & 8.53 \\
V Vuk \etal~\cite{ISPAChallenge} & 6.00 & 1.96 & 2.25 & 0.99 & 18.81 \\
Y Qian \etal~\cite{ISPAChallenge} & 2.21 & 1.32 & 1.41 & 0.43 & 5.65 \\
K Chen \etal~\cite{ISPAChallenge} & 1.84 & 1.27 & 1.32 & 0.39 & \textbf{4.41} \\
Y Qian \etal~\cite{qian2019fast} & 2.27 & 1.26 & 1.35 & 0.39 & 6.02 \\
Afifi \etal 2019~\cite{afifi2019bmvc} & 2.10 & 1.23 & - & 0.47 & 5.38 \\
FFCC~\cite{barron2017fast} (model J) & 2.10 & 1.23 & 1.34 & 0.47 & 5.38 \\
A Savchik \etal~\cite{savchik2019color} & 2.05 & 1.20 & 1.30 & 0.40 & 5.24 \\
WB-sRGB~\cite{Afifi_2019_CVPR,afifi2019bmvc} & \textbf{1.83} & 1.15 & - & \textbf{0.35} & 4.60 \\
\hline
Ours & 1.99 & \textbf{1.06} & \textbf{1.14} & \textbf{0.35} & 5.35 \\
Ours (pretrained) & 1.95 & 1.16 & 1.25 & 0.39 & 4.99 \\
\hline
\end{tabular}
\end{adjustbox}
\caption{Angular error for Cube challenge~\cite{ISPAChallenge}.}
\label{table:cube_test_results}
\end{table}

\subsection{Training on novel sensors}

To explore camera agnostic elements of our model, we train on a combination of the full NUS~\cite{cheng2014illuminant} and Gehler-Shi~\cite{shi2000re,gehler2008bayesian} datasets. As described in \cref{subsec:multidevice_training}, the only remaining device dependent component involves performing illuminant candidate selection per device. 
Once the model is trained, we select candidates from Cube+~\cite{BanicL18a} and test on the Cube challenge dataset~\cite{ISPAChallenge}. We highlight that neither Cube+ nor Cube challenge imagery is seen during model training. 
For meaningful evaluation, we compare against both classical and recent learning-based~\cite{afifi2019bmvc} camera-agnostic methods. Results are shown in \cref{table:cube_test_multidevice}.
We obtain results that are comparable to \cref{table:cube_test_results} without seeing any imagery from our target camera, outperforming both baselines and~\cite{afifi2019bmvc}. \acceptedrev{We clarify that our method performs candidate selection using Cube+~\cite{BanicL18a} to adapt the candidate set to the novel device while~\cite{afifi2019bmvc} does not see any information from the new camera.}

\acceptedrev{We provide additional experimental results for differing values of $K$ ($K$-means candidate selection) in the supplementary material.} We observe stability for $K>=25$. The low number of candidates required is likely linked to the two Cube datasets having reasonably compact distributions.



\begin{table}[b]
\begin{adjustbox}{max width=\columnwidth}
\begin{tabular}{|l|c|c|c|c|c|}
\hline
Method & Mean & Med. & Tri. & Best 25\% & Worst 25\% \\
\hline\hline
Gray-world~\cite{buchsbaum1980spatial} & 4.44 & 3.50 & - & 0.77 & 9.64 \\
1st-order Gray-Edge~\cite{van2007edge} & 3.51 & 2.30 & - & 0.56 & 8.53 \\
Afifi \etal 2019~\cite{afifi2019bmvc} & 2.89 & 1.72 & - & 0.71 & 7.06 \\
\hline
Ours & \textbf{2.07} & \textbf{1.31} & 1.43 & \textbf{0.41} & \textbf{5.12} \\
\hline
\end{tabular}
\end{adjustbox}
\caption{Angular error for the Cube challenge~\cite{ISPAChallenge} trained solely on the dataset of NUS~\cite{cheng2014illuminant} and Gehler-Shi~\cite{shi2000re,gehler2008bayesian}. For our method, candidate selection is performed on Cube+~\cite{BanicL18a} dataset.} 
\label{table:cube_test_multidevice}
\end{table}


\subsection{Qualitative evaluation}

\begin{figure}[b]
\centering
\captionsetup[subfigure]{position=bottom,labelformat=empty}
\subfloat[(a) Input image]{\rule{0.33\columnwidth}{0pt}}
\subfloat[(b) Our prediction]{\rule{0.33\columnwidth}{0pt}}
\subfloat[(c) Ground Truth]{\rule{0.33\columnwidth}{0pt}}
\vspace{-0.25cm}
\subfloat
{
  \centering
  \includegraphics[width=0.30\columnwidth]{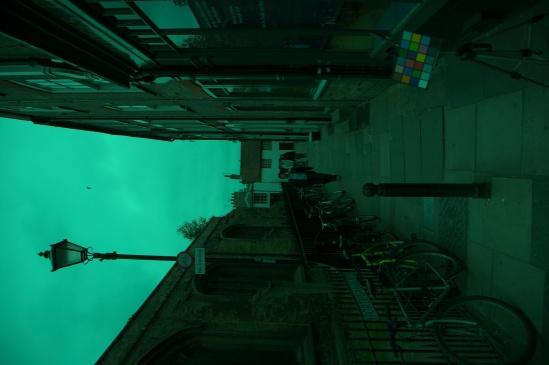}
}
\subfloat[Error: $0.03\degree$]
{
  \centering
  \includegraphics[width=0.03\columnwidth,height=1.66cm]{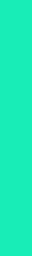}
  \hspace*{-1.5mm}
  \includegraphics[width=0.30\columnwidth]{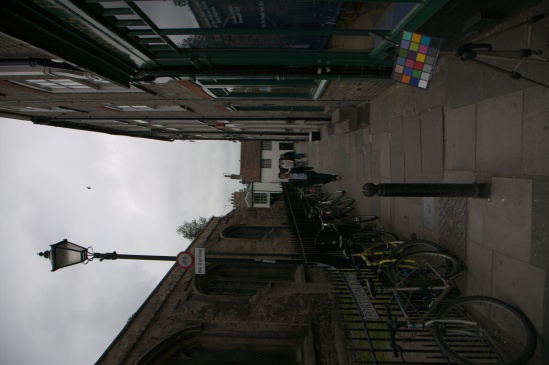}
}
\subfloat
{
  \centering
  \includegraphics[width=0.03\columnwidth,height=1.66cm]{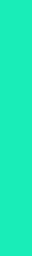}
  \hspace*{-1.5mm}
  \includegraphics[width=0.30\columnwidth]{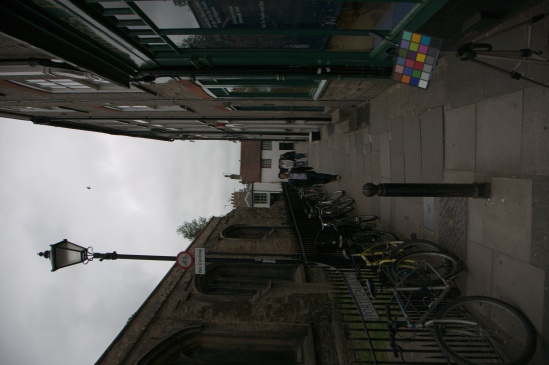}
}
\vspace{-0.25cm}

\subfloat
{
  \centering
  \includegraphics[width=0.30\columnwidth]{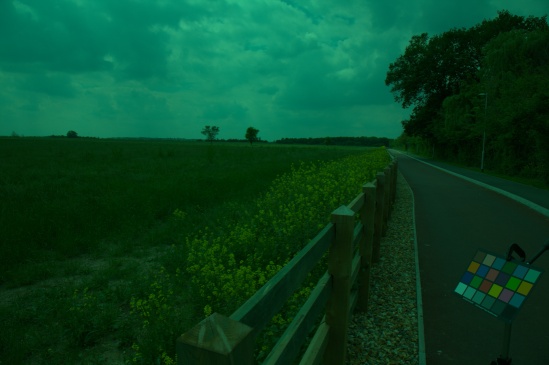}
}
\subfloat[Error: $0.65\degree$]
{
  \centering
  \includegraphics[width=0.03\columnwidth,height=1.66cm]{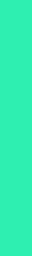}
  \hspace*{-1.5mm}
  \includegraphics[width=0.30\columnwidth]{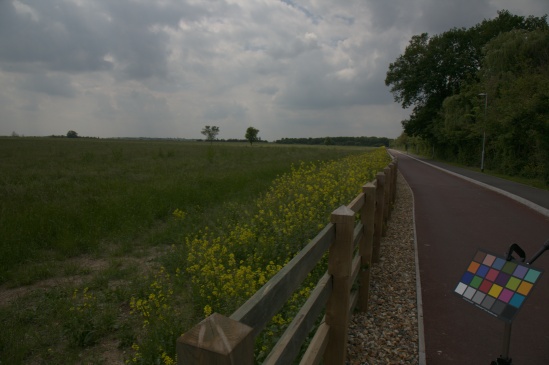}
}
\subfloat
{
  \centering
  \includegraphics[width=0.03\columnwidth,height=1.66cm]{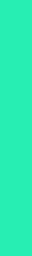}
  \hspace*{-1.5mm}
  \includegraphics[width=0.30\columnwidth]{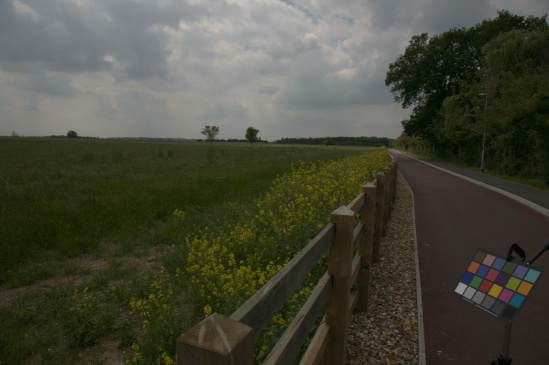}
}
\vspace{-0.25cm}

\subfloat
{
  \centering
  \includegraphics[width=0.30\columnwidth]{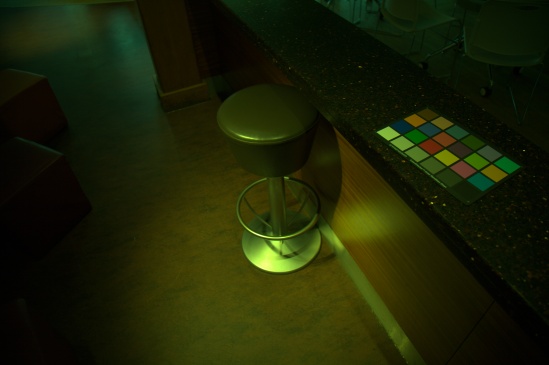}
}
\subfloat[Error: $1.33\degree$]
{
  \centering
  \includegraphics[width=0.03\columnwidth,height=1.66cm]{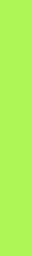}
  \hspace*{-1.5mm}
  \includegraphics[width=0.30\columnwidth]{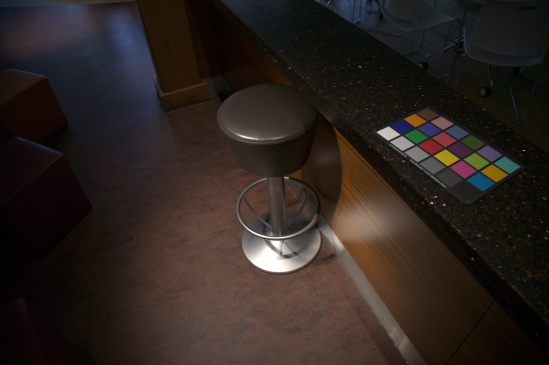}
}
\subfloat
{
  \centering
  \includegraphics[width=0.03\columnwidth,height=1.66cm]{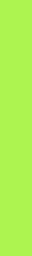}
  \hspace*{-1.5mm}
  \includegraphics[width=0.30\columnwidth]{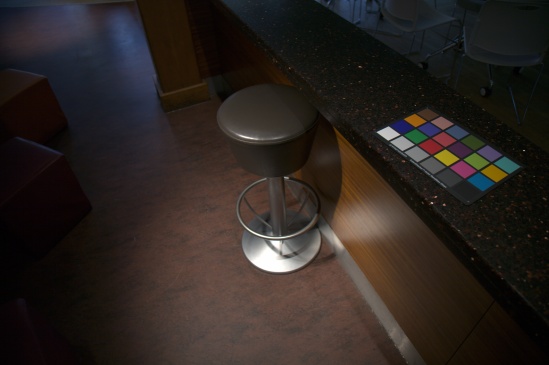}
}
\vspace{-0.25cm}

\subfloat
{
  \centering
  \includegraphics[width=0.30\columnwidth]{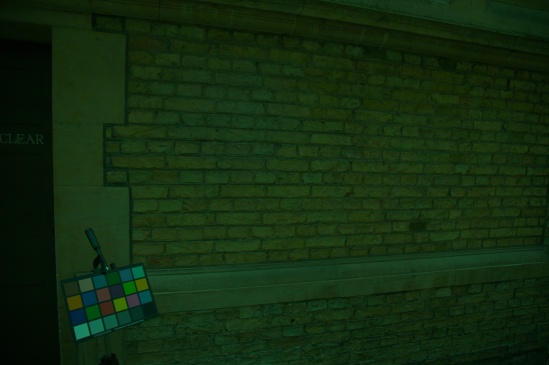}
}
\subfloat[Error: $2.82\degree$]
{
  \centering
  \includegraphics[width=0.03\columnwidth,height=1.66cm]{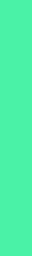}
  \hspace*{-1.5mm}
  \includegraphics[width=0.30\columnwidth]{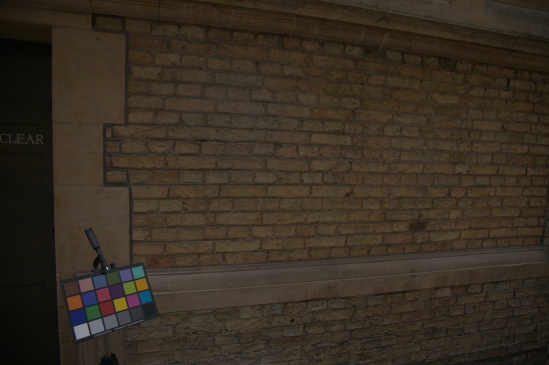}
}
\subfloat
{
  \centering
  \includegraphics[width=0.03\columnwidth,height=1.66cm]{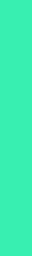}
  \hspace*{-1.5mm}
  \includegraphics[width=0.30\columnwidth]{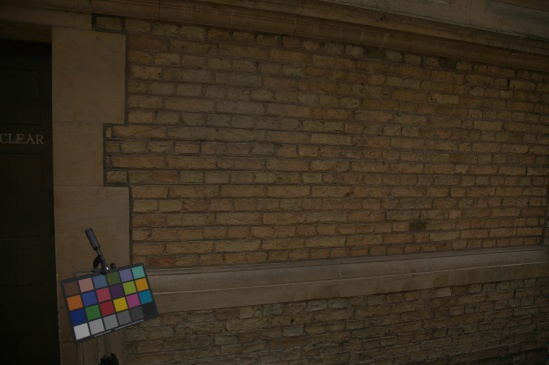}
}
\vspace{-0.25cm}

\subfloat
{
  \centering
  \includegraphics[width=0.30\columnwidth]{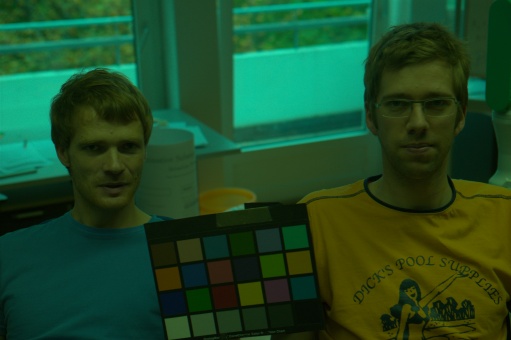}
}
\subfloat[Error: $14.62\degree$]
{
  \centering
  \includegraphics[width=0.03\columnwidth,height=1.66cm]{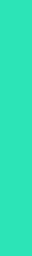}
  \hspace*{-1.5mm}
  \includegraphics[width=0.30\columnwidth]{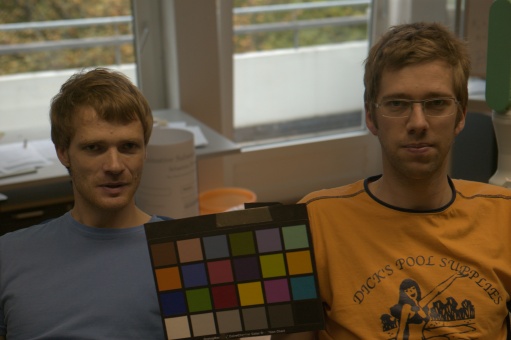}
}
\subfloat
{
  \centering
  \includegraphics[width=0.03\columnwidth,height=1.66cm]{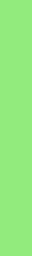}
  \hspace*{-1.5mm}
  \includegraphics[width=0.30\columnwidth]{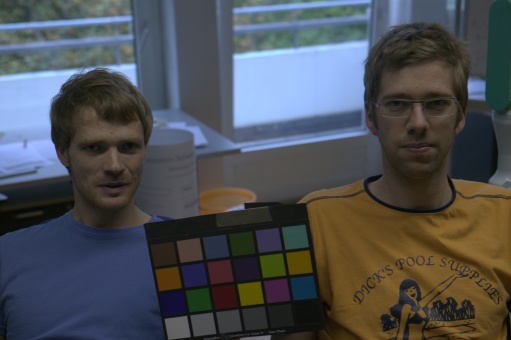}
}
\caption{Example results \acceptedrev{taken} from the Gehler-Shi~\cite{shi2000re,gehler2008bayesian} dataset. Input, our result and ground truth per row. Images to visualise are chosen by sorting all test images using increasing error and evenly sampling images according to that ordering. \acceptedrev{Images are rendered in sRGB color space}.}
\label{fig:visual_results}
\end{figure}

We provide visual results for the Gehler-Shi~\cite{shi2000re,gehler2008bayesian} dataset in \cref{fig:visual_results}. We sort inference results by increasing angular error and sample $5$ images uniformly. For each row, we show (a) the input image (b) our estimated illuminant color and resulting white-balanced image (c) the ground truth illuminant color and resulting white-balanced image. Images are first white-balanced, then, we apply an estimated CCM (Color Correction Matrix), and finally, sRGB gamma correction. We mask out the \emph{Macbeth Color Checker} calibration object during both training and evaluation.

Our most challenging example (\cf last row of \cref{fig:visual_results}) is a multi-illuminant scene (indoor and outdoor lights), we observe our method performs accurate correction for objects illuminated by the outdoor light, yet the ground truth is only measured for the indoor illuminant, hence the high angular error. This highlights the limitation linked to our single global illuminant assumption, common to the majority of CC algorithms. We show additional qualitative results in the supplementary material.

\section{Conclusion}

We propose a novel multi-hypothesis color constancy model capable of effectively learning from image samples that were captured by multiple cameras. We frame the problem under a Bayesian formulation and obtain data-driven likelihood estimates by learning to classify achromatic imagery. We highlight the challenging nature of multi-device learning due to camera color space differences, spectral sensitivity and physical sensor effects. We validate the benefits of our proposed solution for multi-device learning and provide state-of-the-art results on two popular color constancy datasets while maintaining real-time inference constraints. We additionally provide evidence supporting our claims that framing the learning question as a classification task \cf regression can lead to strong performance without requiring model re-training or fine-tuning.

{\small
\bibliographystyle{ieee_fullname}
\bibliography{egbib}
}

\clearpage
\newpage
\mbox{~}

\appendix
\section*{A Multi-Hypothesis Approach to Color Constancy: supplementary material}

\counterwithin{figure}{section}

We provide additional material to supplement our main paper. In~\cref{sec:architecture_details}, we present our shallow CNN architecture. Two experimental studies on the number of illuminant candidates are provided in~\cref{sec:number_candidates}. In~\cref{sec:nus_per_cam}, we report details on NUS~\cite{cheng2014illuminant} per-camera median angular error to provide evidence for our claim that we consistently improve accuracy for each camera, using multi-camera training (see main paper Section $4.4$). In \cref{sec:candidate_selection}, we show additional results from our exploration of candidate selection strategy. \Cref{sec:runtime} provides run-time measurements and in \cref{sec:failure_cases} we observe failure cases and discuss limitations of our method.
Finally, \cref{sec:additional_qualitative} provides additional visual results comparing our method with FFCC~\cite{barron2017fast}.

\section{Architecture details}
\label{sec:architecture_details}

In \cref{table:architecture}, we present our CNN architecture. We propose a shallow CNN, one spatial $3{\times}3$ convolution and two subsequent layers constituting $1{\times}1$ convolutions with a final global spatial pooling. Lastly, three fully connected layers gradually reduce the dimensionality to one.

\begin{table}[h]
\begin{center}
\begin{tabular}{|l|c|c|c|c|c|}
\hline
Layer & Kernel & Input & Output \\
\hline\hline
Conv. & $3{\times}3$ & $64{\times}64{\times}3$ & $64{\times}64{\times}64$ \\
Conv. & $1{\times}1$ & $64{\times}64{\times}64$ & $64{\times}64{\times}64$ \\
Conv. & $1{\times}1$ & $64{\times}64{\times}64$ & $64{\times}64{\times}128$ \\
Avg. Pool. & $64{\times}64$ & $64{\times}64{\times}128$ & $128$ \\
FC & - & $128$ & $64$ \\
FC & - & $64$ & $32$ \\
FC & - & $32$ & $1$ \\
\hline
\end{tabular}
\end{center}
\caption{CNN architecture details. Fully connected layers and convolutions are followed by a ReLU activation except the last layer.}
\label{table:architecture}
\end{table}

\section{Number of illuminant candidates}
\label{sec:number_candidates}

In \cref{table:candidates} we present a study varying the number of candidate illuminants produced by $K$-means. We find experimentally that accuracy improves with the number of cluster centres until a plateau is reached, suggesting that we need ${\sim}100$ candidate illuminants to achieve competitive angular error for the Gehler-Shi dataset~\cite{shi2000re,gehler2008bayesian}.

\begin{table}[t]
\begin{adjustbox}{max width=\columnwidth}
\begin{tabular}{|l|c|c|c|c|c|}
\hline
\# candidates & Mean & Med. & Tri. & Best 25\% & Worst 25\% \\
\hline\hline
5 & 2.79 & 2.06 & 2.20 & 0.67 & 6.23 \\
25 & 2.24 & 1.50 & 1.64 & 0.38 & 7.34 \\
50 & 2.25 & 1.47 & 1.66 & 0.37 & 5.51 \\
100 & 2.15 & 1.38 & 1.55 & 0.40 & 5.16 \\
120 & 2.10 & 1.32 & 1.53 & 0.36 & 5.10 \\
150 & 2.16 & 1.33 & 1.53 & 0.39 & 5.25 \\
200 & 2.16 & 1.39 & 1.59 & 0.37 & 5.20 \\
\hline
\end{tabular}
\end{adjustbox}
\caption{Error for differing number of candidates for $K$-means candidate selection. Angular error for Gehler-Shi dataset~\cite{shi2000re,gehler2008bayesian}.}
\label{table:candidates}
\end{table}

Additionally, we provide analogous results for different values of $K$ for $K$-means candidate selection for the training-free model (see main paper Section 4.5), in~\cref{table:cube_test_multidevice_k}. We observe stability for $K>=25$. The low number of candidates required is likely linked to the two Cube datasets having reasonably compact illuminant distributions.

\begin{table}[t]
\begin{adjustbox}{max width=\columnwidth}
\begin{tabular}{|l|c|c|c|c|c|}
\hline
\# candidates & Mean & Med. & Tri. & Best 25\% & Worst 25\% \\
\hline\hline
5 & 2.53 & 1.71 & 1.81 & 0.51 & 6.06 \\
25 & 2.28 & 1.43 & 1.59 & 0.45 & 5.63 \\
50 & 2.28 & 1.46 & 1.61 & 0.46 & 5.52 \\
100 & 2.12 & 1.31 & 1.45 & 0.40 & 5.31 \\
120 & 2.07 & 1.31 & 1.43 & 0.41 & 5.12 \\
150 & 2.16 & 1.32 & 1.49 & 0.40 & 5.34 \\
200 & 2.12 & 1.33 & 1.47 & 0.40 & 5.27 \\
\hline
\end{tabular}
\end{adjustbox}
\caption{Angular error for the Cube challenge~\cite{ISPAChallenge} trained only on NUS~\cite{cheng2014illuminant} and Gehler-Shi~\cite{shi2000re,gehler2008bayesian}. For our method, candidate selection is performed on Cube+~\cite{BanicL18a} with varying $K$ for $K$-means candidate selection.}
\label{table:cube_test_multidevice_k}
\end{table}

\section{NUS per-camera median angular error}
\label{sec:nus_per_cam}

We provide evidence supporting our paper claim that training the proposed model with images from multiple cameras outperforms individual, per-camera, model training (see Section $4.4$, of the main paper).



We reiterate that
folds are divided such that scene content is consistent within a fold, across all cameras. This ensures to avoid testing on familiar scene content, as observed by a different camera during training. \acceptedrev{Towards reproducibility, and fair comparison, our suppplementary material provides the cross validation (CV) splits, used in the main paper, for multi-device training. CV splits were generated manually by ensuring that all images of the same scene (across different cameras) belong to the same fold.}

In~\cref{table:nus8_camera_results} we report median angular-error for test images of the NUS~\cite{cheng2014illuminant} dataset. 
Multi-device training can be seen to consistently improve the median angular error for all NUS cameras at test time.

\begin{table}[t]
\begin{adjustbox}{max width=\columnwidth}
\begin{tabular}{|l|c|c|c|c|c|}
\hline
Camera & Ours (one model per device) & Ours (multi-device training) \\
\hline\hline
Canon EOS-1Ds Mark III & 1.59 & 1.49 \\
Canon EOS 600D & 1.49 & 1.23 \\
Fujifilm X-M1 & 1.34 & 1.33 \\
Nikon D5200 & 1.69 & 1.50 \\
Olympus E-PL6 & 1.30 & 1.13 \\
Panasonic Lumix DMC-GX1 & 1.43 & 1.21 \\
Samgsung NX2000 & 1.54 & 1.42 \\
Sony SLT-A57 & 1.50 & 1.41 \\
\hline
\end{tabular}
\end{adjustbox}
\caption{Median angular error of our method for each individual camera of NUS~\cite{cheng2014illuminant}.}
\label{table:nus8_camera_results}
\end{table}

\section{Candidate selection methods}
\label{sec:candidate_selection}

We report additional illuminant candidate selection strategies explored during our investigation.

\noindent \textbf{Uniform-sampling}: we consider the global extrema of our measured illuminant samples (max. and min. in each color space dimension) and sample $n$ points uniformly using an [$\frac{r}{g}$, $\frac{b}{g}$] color space. These samples constitute our illuminant candidates.

\noindent \textbf{$K$-means clustering}: cluster centroids define candidates, as detailed in the main paper, Section $3.2$ and other recent color constancy work~\cite{OhK17}. \acceptedrev{We use RGB color space for clustering, and experimentally verified that both [$\frac{r}{g}$, $\frac{b}{g}$] and RGB color spaces provided similar accuracy.}

\noindent \textbf{Mixture Model (GMM)}: we fit a GMM to our measured illuminant samples in [$\frac{r}{g}$, $\frac{b}{g}$] color space, and then draw $n$ samples from the GMM to define illuminant candidates.

We use $121$ candidates ($11{\times}11$ grid) for uniform candidate selection. For GMM candidate selection, we fit $10$ two-dimensional Gaussian distributions and sample $120$ candidates.

\begin{table}[b]
\begin{adjustbox}{max width=\columnwidth}
\begin{tabular}{|l|c|c|c|c|c|}
\hline
Method & Mean & Med. & Tri. & Best 25\% & Worst 25\% \\
\hline\hline
Uniform & 2.11 & 1.20 & 1.30 & 0.41 & 5.45 \\
GMM & 2.27 & 1.10 & 1.25 & 0.41 & 6.31 \\
$K$-means & 1.99 & 1.06 & 1.14 & 0.35 & 5.35 \\
\hline
\end{tabular}
\end{adjustbox}
\caption{Angular error for Cube challenge~\cite{ISPAChallenge} of our method using different candidate selection methods.}
\label{table:candidate_selection}
\end{table}

In~\cref{table:candidate_selection} we report inference performance on the Cube challenge~\cite{ISPAChallenge} data set using the described candidate selection strategies. We observe that simple uniform-sampling candidate selection performs reasonably well. The strategy provides an extremely simple implementation yet, by definition, will also sample some portion of very unlikely candidates. We note, however, that if the interpolation between candidates span the illuminant space, our method can learn to interpolate these candidates appropriately, accounting for this. The GMM approach also results in slightly weaker accuracy performance \cf $K$-means, motivating our choice of sampling strategy in the experimental work for the main paper.

\section{Inference run-time}
\label{sec:runtime}

We report inference run-time results for the Gehler-Shi dataset~\cite{shi2000re,gehler2008bayesian} in \cref{table:gehler_shi_runtime}. We note that our real-time inference speed is obtained using a Nvidia \emph{Tesla V100} card and unoptimised implementation (PyTorch 1.0~\cite{steiner2019pytorch}). \acceptedrev{We highlight that our algorithm is highly parallelizable, each illuminant candidate likelihood can be computed independently, however, we obtain the run-time with single-thread implementation.} Our input image resolution is $64{\times}64$ and timing results are recorded using $K$-means candidate selection with $K{=}120$. The timing performance of other methods are obtained from their respective citations. \acceptedrev{We acknowledge that timing comparisons are non-rigorous; reported run-times are measured using differing hardware. To provide additional fair comparison; \cref{table:gehler_shi_runtime_fair_comparison} reports run-times for both our method and the official\footnote{https://github.com/google/ffcc} FFCC~\cite{barron2017fast} implementation run on Matlab R2019b, under common hardware (Intel Core i9-9900X (3.50GHz)).}

\begin{table}[ht]
\begin{adjustbox}{max width=\columnwidth}
\begin{tabular}{|l|c|c|}
\hline
Method & Run-time (ms) & Hardware \\
\hline\hline
CCC~\cite{barron2015convolutional} & 520 & 2012 HP Z420 workstation (CPU) \\
Cheng \etal 2015~\cite{cheng2015effective} & 250 & Intel Xeon 3.5GHz (CPU) \\
FC4~\cite{hu2017fc4} & 25 & Nvidia GTX TITAN X Maxwell (GPU) \\
FFCC~\cite{barron2017fast} (model Q) & 1.1 &  Intel Xeon CPU E5-2680 (CPU) \\
CM 2019~\cite{gong2019bmvc} & 1 & Nvidia Tesla K40m (GPU) \\
\hline
Ours & 7.3 & Nvidia Tesla V100 (GPU) \\
\hline
\end{tabular}
\end{adjustbox}
\caption{Inference time for images of Gehler-Shi dataset~\cite{shi2000re,gehler2008bayesian}. Run-time is provided in milliseconds (ms).}
\label{table:gehler_shi_runtime}
\end{table}

\begin{table}[ht]
\begin{adjustbox}{max width=\columnwidth}
\begin{tabular}{|l|c|c|}
\hline
Method & Run-time (ms)\\
\hline\hline
FFCC~\cite{barron2017fast} (model Q) & 1.2\\
Ours & 128\\
\hline
\end{tabular}
\end{adjustbox}
\caption{Inference time for images of Gehler-Shi dataset~\cite{shi2000re,gehler2008bayesian}. Run-time is provided in milliseconds (ms). Run-time measured using a Intel Core i9-9900X (3.50GHz) CPU.}
\label{table:gehler_shi_runtime_fair_comparison}
\end{table}

\section{Failure cases}
\label{sec:failure_cases}

\begin{figure}[h]
\centering
\captionsetup[subfigure]{position=bottom,labelformat=empty}

\subfloat[(a) Our prediction (angular-error~{=}~$20.12\degree$)]
{
  \centering
  \includegraphics[width=0.03\columnwidth,height=5.65cm]{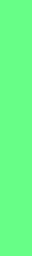}
  \hspace*{-1.5mm}
  \includegraphics[width=0.45\columnwidth]{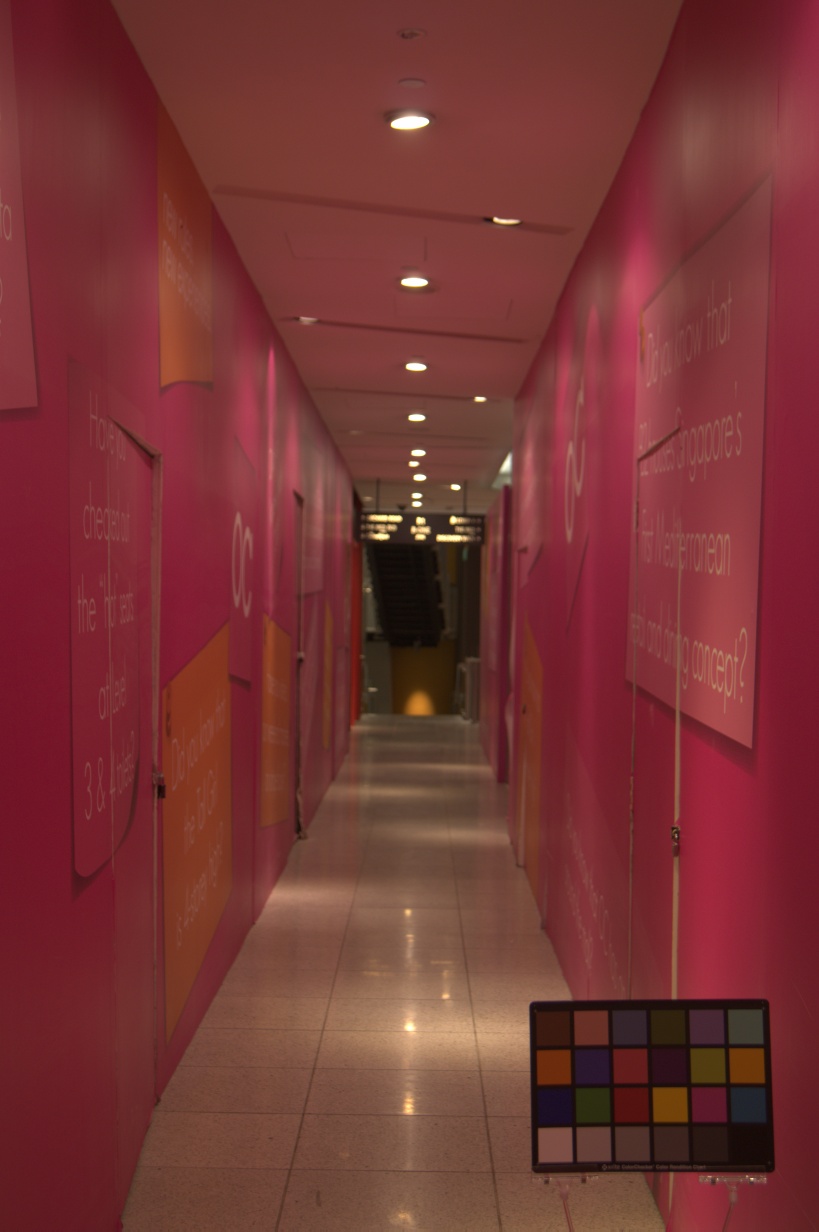}
  \label{fig:failure_case1:a}
}
\subfloat[(b) Ground Truth]
{
  \centering
  \includegraphics[width=0.03\columnwidth,height=5.65cm]{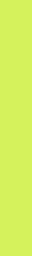}
  \hspace*{-1.5mm}
  \includegraphics[width=0.45\columnwidth]{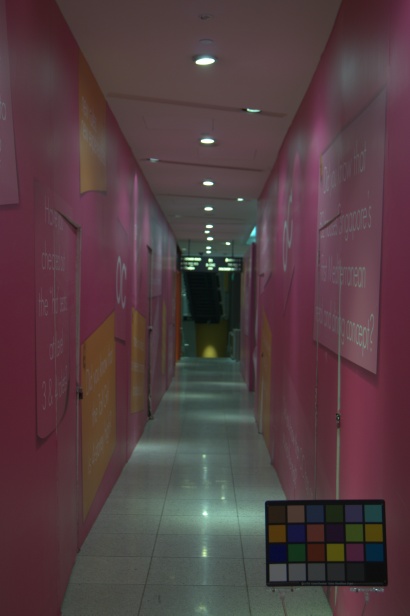}
  \label{fig:failure_case1:b}
}

\subfloat[(c) $\frac{r}{g}$, $\frac{b}{g}$ plot of candidates]
{
  \centering
  \includegraphics[width=0.95\columnwidth]{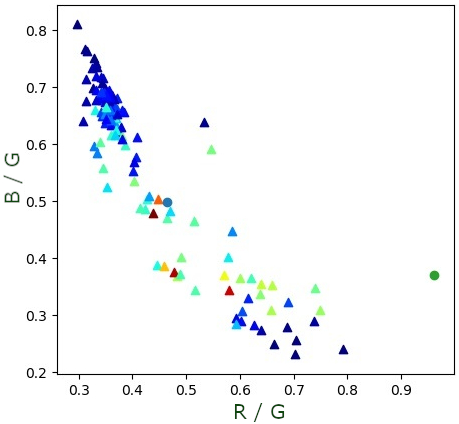}
  \label{fig:failure_case1:c}
}
\caption{This challenging scene is illuminated by a measured illumination color not seen during training. In \cref{fig:failure_case1:c} the green circular point corresponds to the ground-truth illuminant and can be observed to be outwith the illuminant candidate distribution. \acceptedrev{Images are rendered in sRGB color space.}}
\label{fig:failure_case1}
\end{figure}

\begin{figure}[h]
\centering
\captionsetup[subfigure]{position=bottom,labelformat=empty}

\subfloat[(a) Our prediction (angular-error~{=}~$6.14\degree$)]
{
  \centering
  \includegraphics[width=0.03\columnwidth,height=5.65cm]{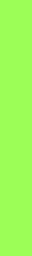}
  \hspace*{-1.5mm}
  \includegraphics[width=0.45\columnwidth]{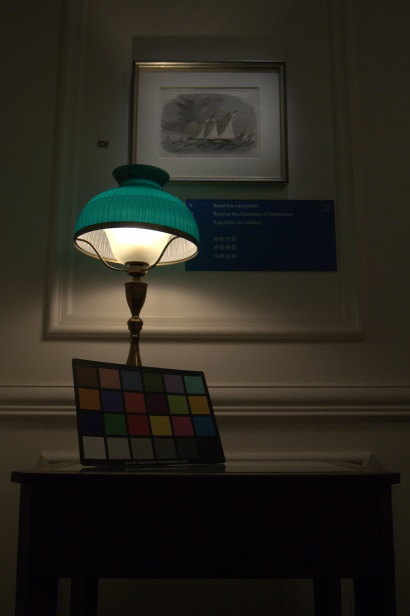}
}
\subfloat[(b) Ground Truth]
{
  \centering
  \includegraphics[width=0.03\columnwidth,height=5.65cm]{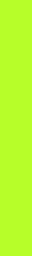}
  \hspace*{-1.5mm}
  \includegraphics[width=0.45\columnwidth]{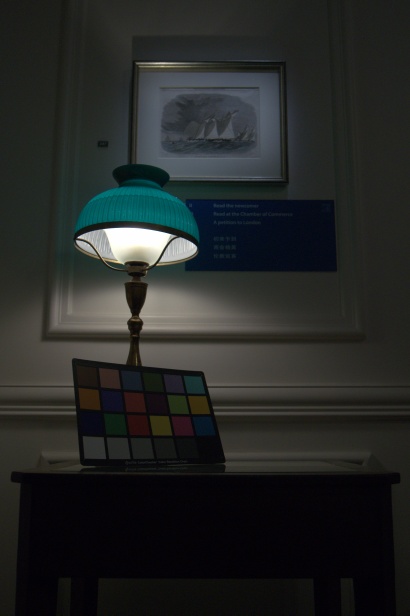}
}
\caption{This scene can be observed to be illuminated by more than one light source, breaking the single global illuminant assumption. \acceptedrev{Images are rendered in sRGB color space.}}
\label{fig:failure_case2}
\end{figure}

\begin{figure}[t]
\centering
\captionsetup[subfigure]{position=bottom,labelformat=empty}

\subfloat[(a) Our prediction (angular-error~{=}~$6.05\degree$)]
{
  \centering
  \includegraphics[width=0.03\columnwidth,height=5.65cm]{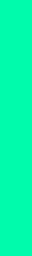}
  \hspace*{-1.5mm}
  \includegraphics[width=0.45\columnwidth]{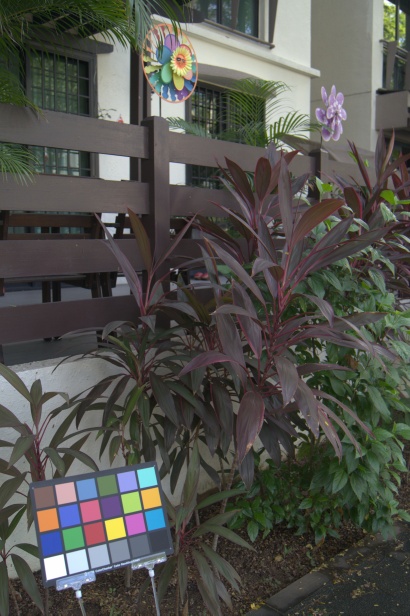}
}
\subfloat[(b) Ground Truth]
{
  \centering
  \includegraphics[width=0.03\columnwidth,height=5.65cm]{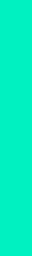}
  \hspace*{-1.5mm}
  \includegraphics[width=0.45\columnwidth]{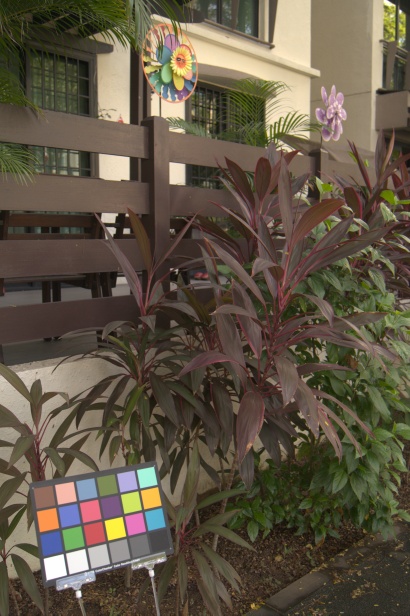}
}
\caption{An ambiguous scene with multiple \emph{plausible} solutions, highlighting the ill-posed nature of the color constancy problem. Our method infers a plausible, yet incorrect, solution; that the color of the stone building is white. \acceptedrev{Images are rendered in sRGB color space.}}
\label{fig:failure_case3}
\end{figure}

In \cref{fig:failure_case1,fig:failure_case2,fig:failure_case3} we provide observed limitations and failure cases. Our method learns to interpolate between candidate illuminants, that are observed during training, but not to extrapolate to new illuminants. In \cref{fig:failure_case1:c}, the ground truth illuminant (green filled circle) is clearly out of distribution, with no similar candidate illuminants observed during training. The resulting inference accuracy in \cref{fig:failure_case1:a} suffers as a result.

Further, our single global illuminant assumption can be seen to be violated in~\cref{fig:failure_case2}. The predicted illuminant attempts to balance the outer boundary portions of the wall painting as achromatic, clearly illuminated from above (out of shot). The measured ground truth illuminant captures the desk lamp illumination, resulting in high angular error for this image due to the global assumption.

Finally, in~\cref{fig:failure_case3}, we observe an example scene with extreme ambiguities. Our method appears to infer that the stone building in the scene background is achromatic, producing a highly plausible image. Yet the measured ground-truth illuminant illustrates the true building color to be of mild beige-yellow.

\section{Additional qualitative results}
\label{sec:additional_qualitative}

\begin{figure*}[ht]
\centering
\captionsetup[subfigure]{position=bottom,labelformat=empty}

\subfloat[(a) Input image]
{
  \centering
  \includegraphics[width=0.22\textwidth]{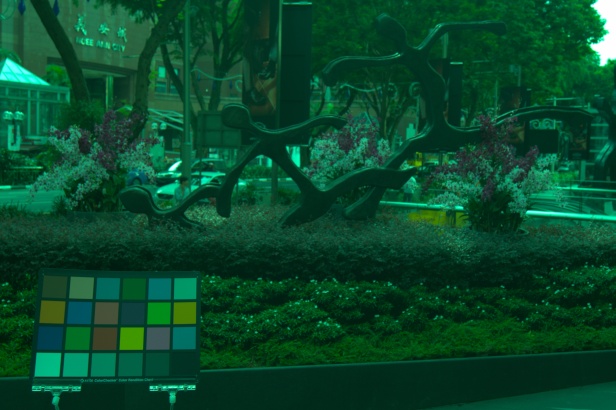}
}
\subfloat[(b) FFCC (error: $0.08\degree$)]
{
  \centering
  \includegraphics[width=0.03\columnwidth,height=2.55cm]{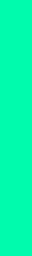}
  \hspace*{-1.5mm}
  \includegraphics[width=0.22\textwidth]{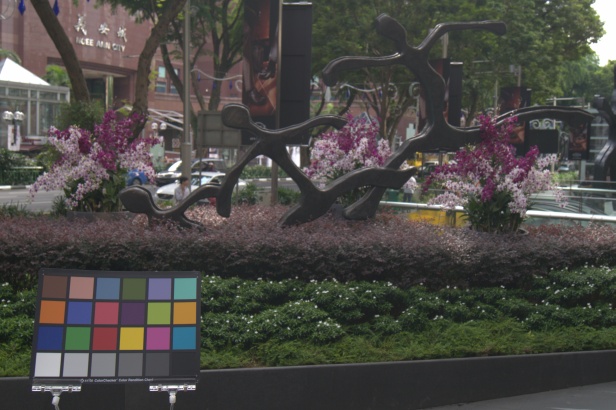}
}
\subfloat[(c) Ours (error: $0.07\degree$)]
{
  \centering
  \includegraphics[width=0.03\columnwidth,height=2.55cm]{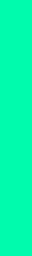}
  \hspace*{-1.5mm}
  \includegraphics[width=0.22\textwidth]{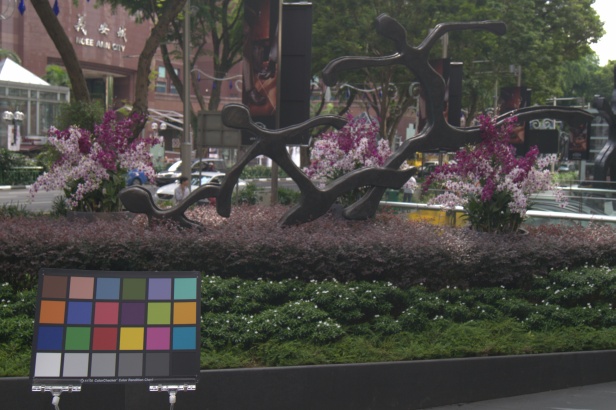}
}
\subfloat[(d) Ground Truth]
{
  \centering
  \includegraphics[width=0.03\columnwidth,height=2.55cm]{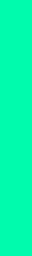}
  \hspace*{-1.5mm}
  \includegraphics[width=0.22\textwidth]{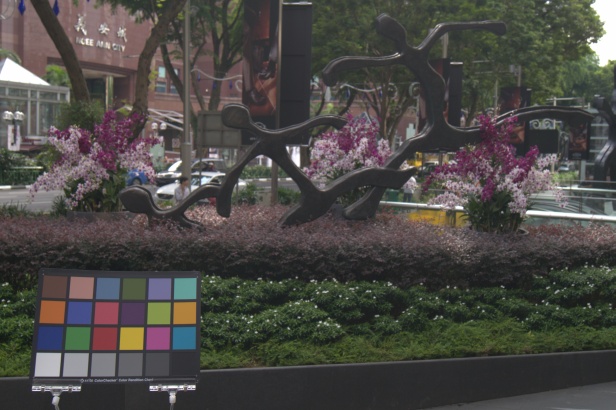}
}

\subfloat[(a) Input image]
{
  \centering
  \includegraphics[width=0.22\textwidth]{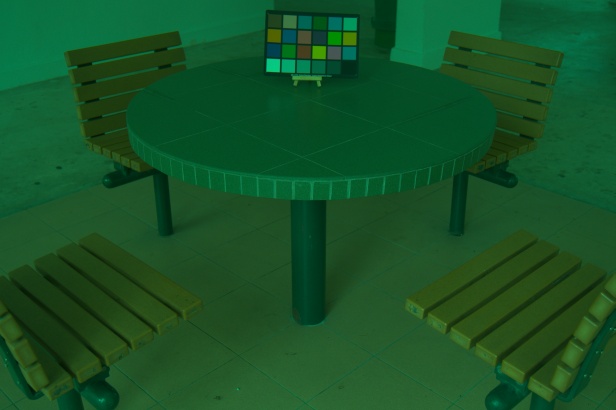}
}
\subfloat[(b) FFCC (error: $1.66\degree$)]
{
  \centering
  \includegraphics[width=0.03\columnwidth,height=2.55cm]{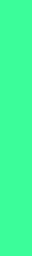}
  \hspace*{-1.5mm}
  \includegraphics[width=0.22\textwidth]{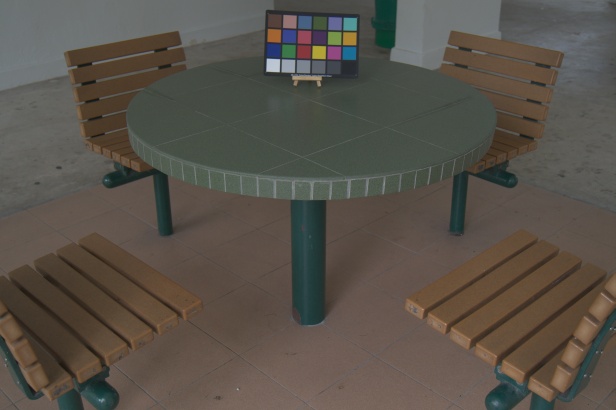}
}
\subfloat[(c) Ours (error: $0.18\degree$)]
{
  \centering
  \includegraphics[width=0.03\columnwidth,height=2.55cm]{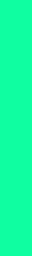}
  \hspace*{-1.5mm}
  \includegraphics[width=0.22\textwidth]{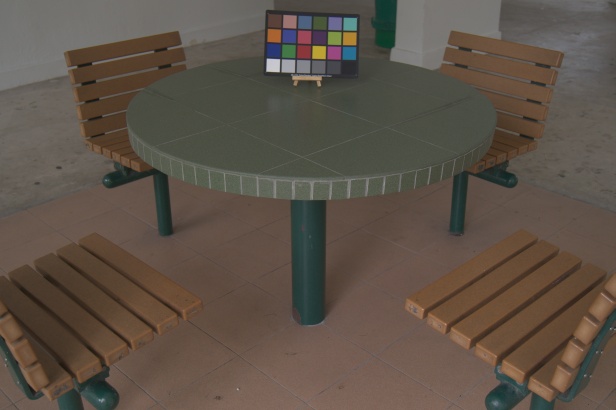}
}
\subfloat[(d) Ground Truth]
{
  \centering
  \includegraphics[width=0.03\columnwidth,height=2.55cm]{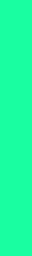}
  \hspace*{-1.5mm}
  \includegraphics[width=0.22\textwidth]{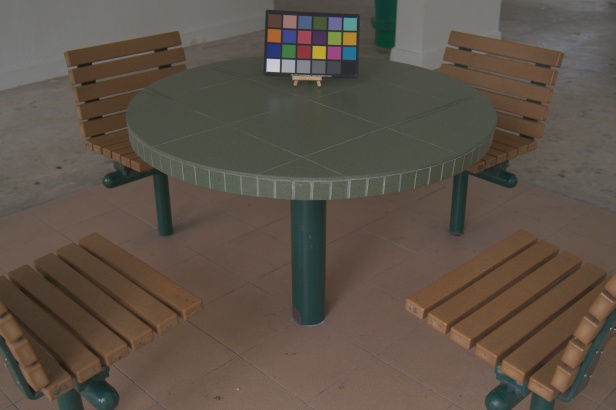}
}

\subfloat[(a) Input image]
{
  \centering
  \includegraphics[width=0.22\textwidth]{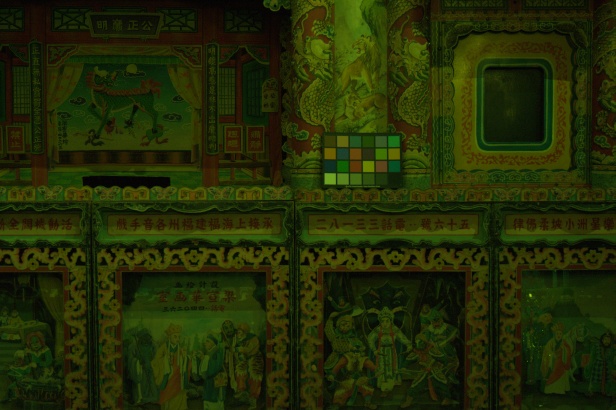}
}
\subfloat[(b) FFCC (error: $1.66\degree$)]
{
  \centering
  \includegraphics[width=0.03\columnwidth,height=2.55cm]{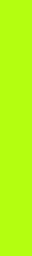}
  \hspace*{-1.5mm}
  \includegraphics[width=0.22\textwidth]{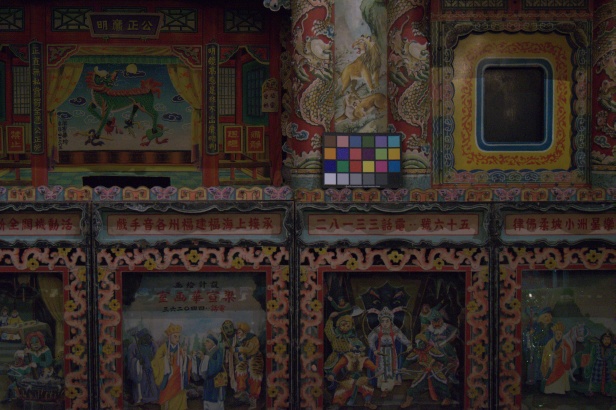}
}
\subfloat[(c) Ours (error: $1.41\degree$)]
{
  \centering
  \includegraphics[width=0.03\columnwidth,height=2.55cm]{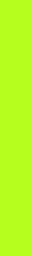}
  \hspace*{-1.5mm}
  \includegraphics[width=0.22\textwidth]{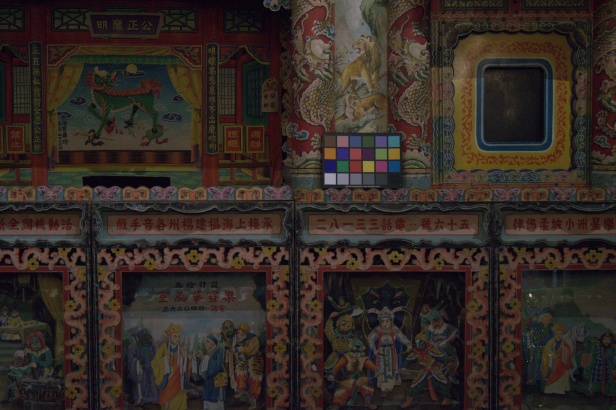}
}
\subfloat[(d) Ground Truth]
{
  \centering
  \includegraphics[width=0.03\columnwidth,height=2.55cm]{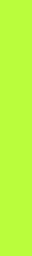}
  \hspace*{-1.5mm}
  \includegraphics[width=0.22\textwidth]{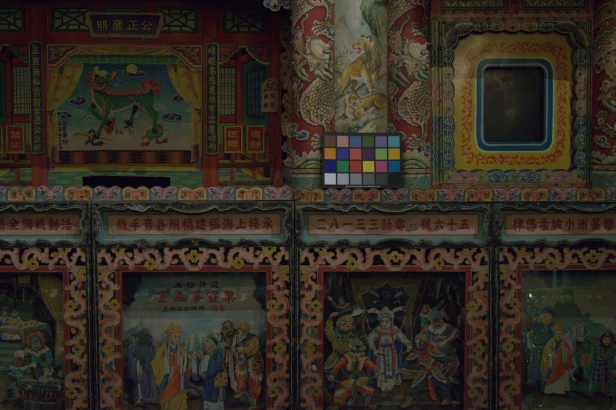}
}

\subfloat[(a) Input image]
{
  \centering
  \includegraphics[width=0.22\textwidth]{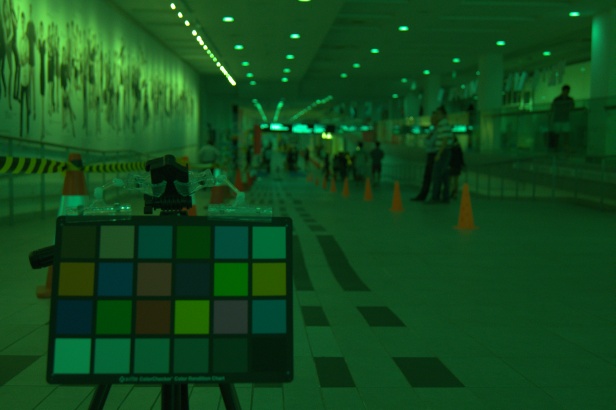}
}
\subfloat[(b) FFCC (error: $2.68\degree$)]
{
  \centering
  \includegraphics[width=0.03\columnwidth,height=2.55cm]{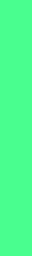}
  \hspace*{-1.5mm}
  \includegraphics[width=0.22\textwidth]{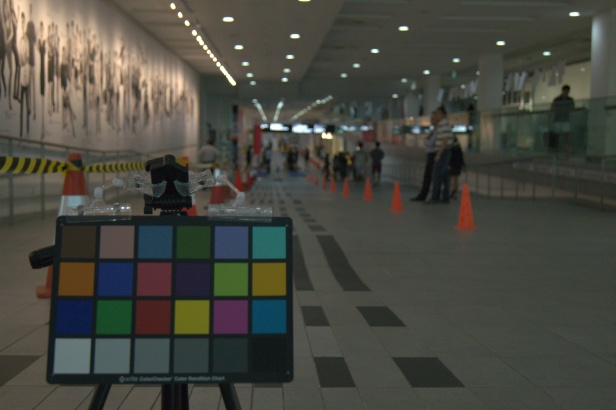}
}
\subfloat[(c) Ours (error: $2.50\degree$)]
{
  \centering
  \includegraphics[width=0.03\columnwidth,height=2.55cm]{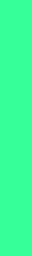}
  \hspace*{-1.5mm}
  \includegraphics[width=0.22\textwidth]{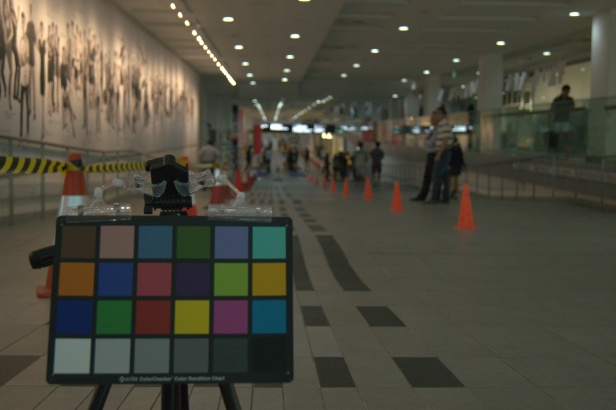}
}
\subfloat[(d) Ground Truth]
{
  \centering
  \includegraphics[width=0.03\columnwidth,height=2.55cm]{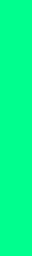}
  \hspace*{-1.5mm}
  \includegraphics[width=0.22\textwidth]{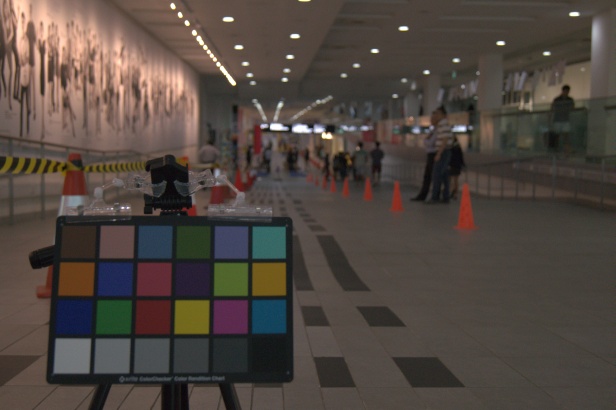}
}

\subfloat[(a) Input image]
{
  \centering
  \includegraphics[width=0.22\textwidth]{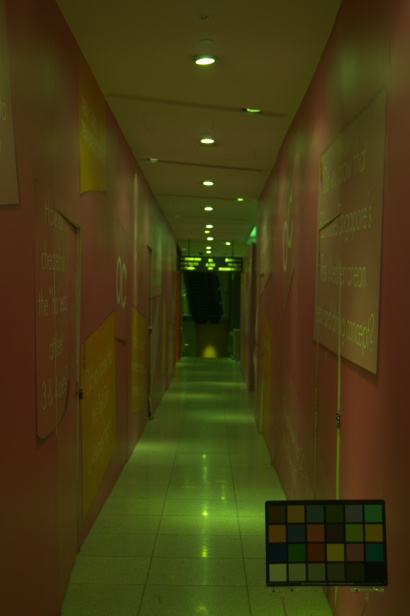}
}
\subfloat[(b) FFCC (error: $11.93\degree$)]
{
  \centering
  \includegraphics[width=0.03\columnwidth,height=5.75cm]{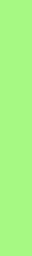}
  \hspace*{-1.5mm}
  \includegraphics[width=0.22\textwidth]{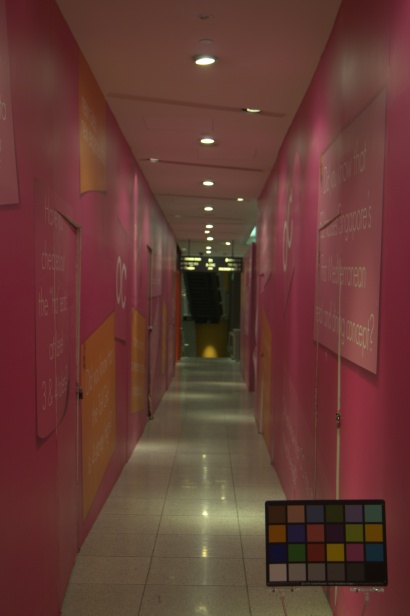}
}
\subfloat[(c) Ours (error: $20.12\degree$)]
{
  \centering
  \includegraphics[width=0.03\columnwidth,height=5.75cm]{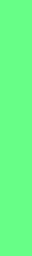}
  \hspace*{-1.5mm}
  \includegraphics[width=0.22\textwidth]{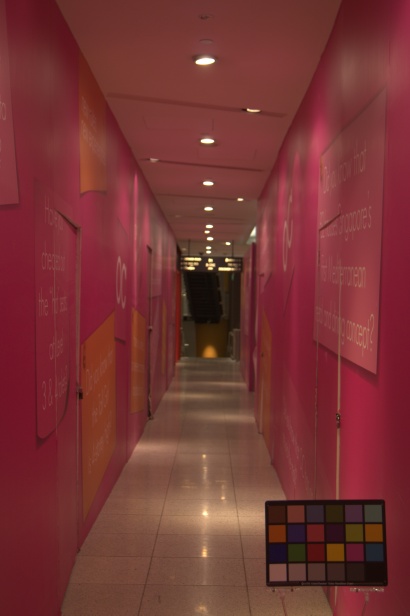}
}
\subfloat[(d) Ground Truth]
{
  \centering
  \includegraphics[width=0.03\columnwidth,height=5.75cm]{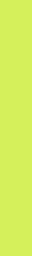}
  \hspace*{-1.5mm}
  \includegraphics[width=0.22\textwidth]{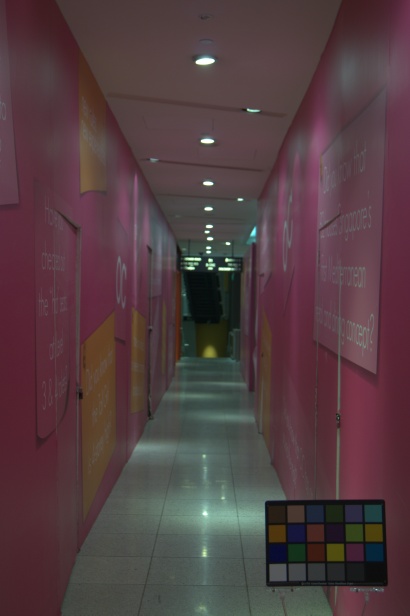}
}
\caption{\acceptedrev{Visual comparisons of FFCC~\cite{barron2017fast} and our method. We sort test results of the Sony dataset (NUS~\cite{cheng2014illuminant}) by the combined (sum total) mean angular error of the two evaluated methods and then uniformly sample images to select test images. Images are rendered in sRGB color space.}}
\label{fig:extended_visual_results}
\end{figure*}

In~\cref{fig:extended_visual_results}, we provide additional qualitative results in the form of test images from the NUS~\cite{cheng2014illuminant} dataset (Sony camera). For each test sample we show the input image and a white-balanced image, corrected using the ground-truth illumination in addition to the output of our model (``multi-device training + pretraining''), and that of FFCC (model Q)~\cite{barron2017fast}. Each row consists of: (a) the input image (b) FFCC~\cite{barron2017fast} (c) our prediction (d) ground truth.

In similar fashion to~\cite{barron2015convolutional}, we adopt the strategy of sorting test images by the combined mean angular-error of the two evaluated methods. We present images of increasing average difficulty, sampled with a uniform spacing. Images are corrected by inferred illuminants, applying an estimated CCM (Color Correction Matrix), and standard sRGB gamma correction. The \emph{Macbeth Color Checker} is used to generate the ground-truth and is present in the images, however the relevant regions are masked during both training and inference. It can be observed in~\cref{fig:extended_visual_results} in almost all sampled cases, we see consistently improved results with our approach.

\begin{figure*}[ht]
\centering
\captionsetup[subfigure]{position=bottom,labelformat=empty}

\subfloat[(a) Input image]
{
  \centering
  \includegraphics[width=0.22\textwidth]{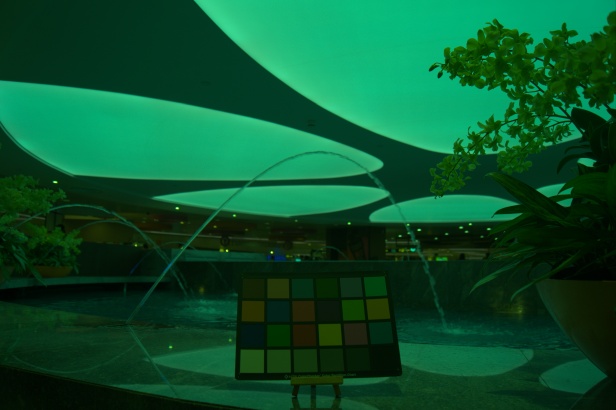}
}
\subfloat[(b) FFCC (error: $11.87\degree$)]
{
  \centering
  \includegraphics[width=0.03\columnwidth,height=2.55cm]{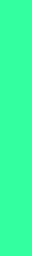}
  \hspace*{-1.5mm}
  \includegraphics[width=0.22\textwidth]{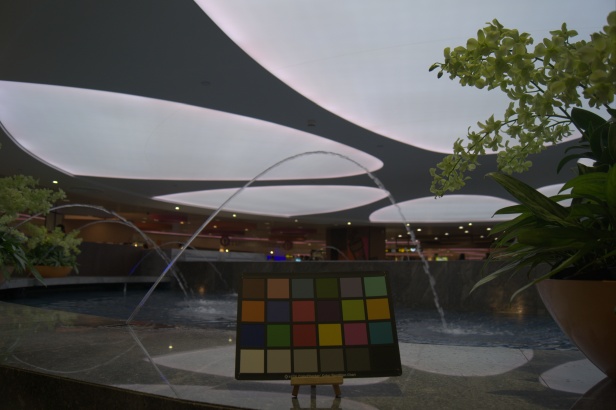}
}
\subfloat[(c) Ours (error: $13.17\degree$)]
{
  \centering
  \includegraphics[width=0.03\columnwidth,height=2.55cm]{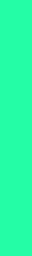}
  \hspace*{-1.5mm}
  \includegraphics[width=0.22\textwidth]{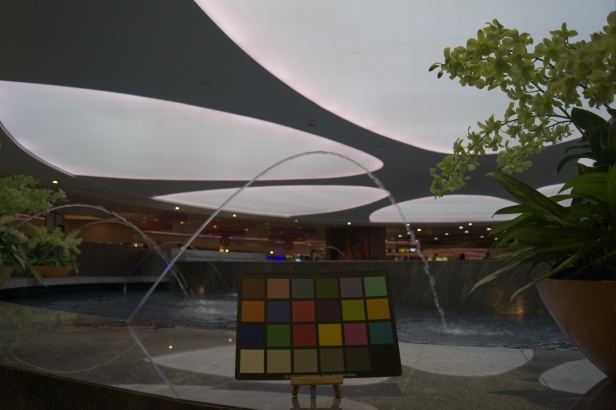}
}
\subfloat[(d) Ground Truth]
{
  \centering
  \includegraphics[width=0.03\columnwidth,height=2.55cm]{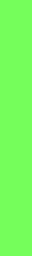}
  \hspace*{-1.5mm}
  \includegraphics[width=0.22\textwidth]{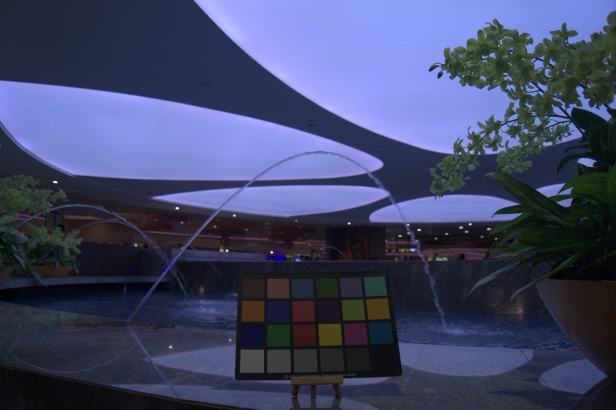}
}

\subfloat[(a) Input image]
{
  \centering
  \includegraphics[width=0.22\textwidth]{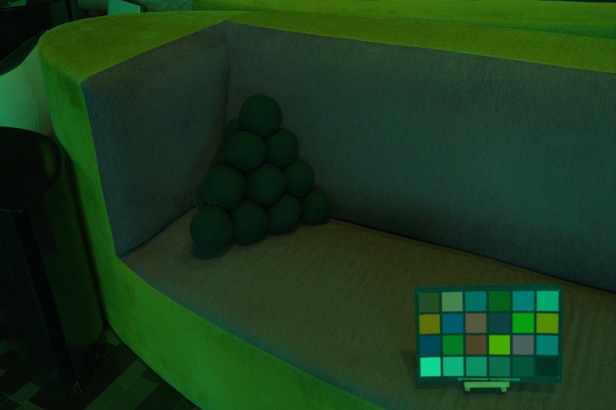}
}
\subfloat[(b) FFCC (error: $16.66\degree$)]
{
  \centering
  \includegraphics[width=0.03\columnwidth,height=2.55cm]{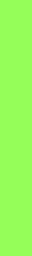}
  \hspace*{-1.5mm}
  \includegraphics[width=0.22\textwidth]{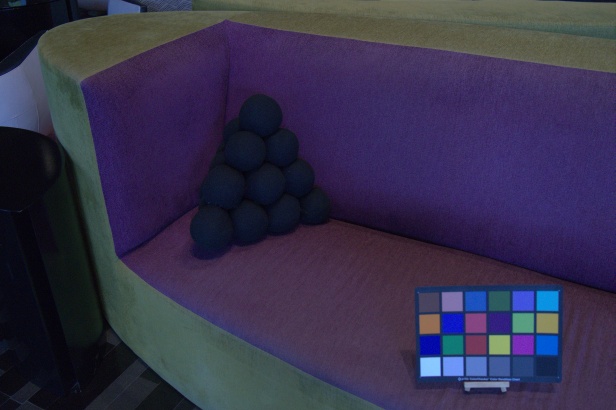}
}
\subfloat[(c) Ours (error: $7.24\degree$)]
{
  \centering
  \includegraphics[width=0.03\columnwidth,height=2.55cm]{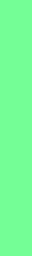}
  \hspace*{-1.5mm}
  \includegraphics[width=0.22\textwidth]{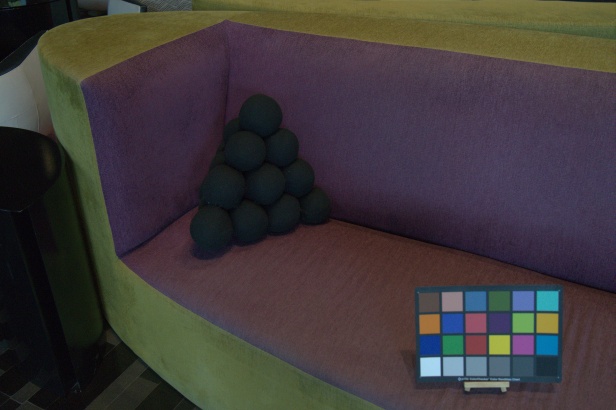}
}
\subfloat[(d) Ground Truth]
{
  \centering
  \includegraphics[width=0.03\columnwidth,height=2.55cm]{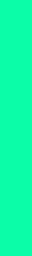}
  \hspace*{-1.5mm}
  \includegraphics[width=0.22\textwidth]{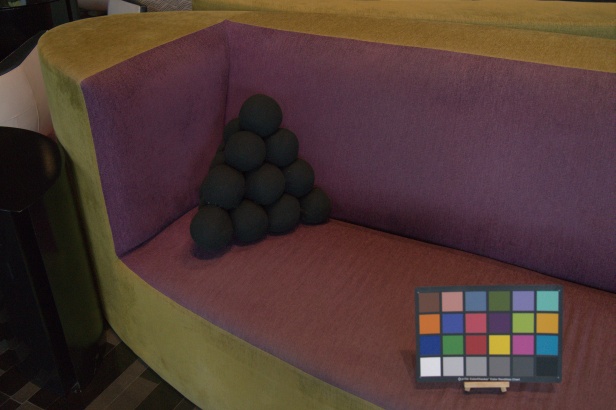}
}

\subfloat[(a) Input image]
{
  \centering
  \includegraphics[width=0.22\textwidth]{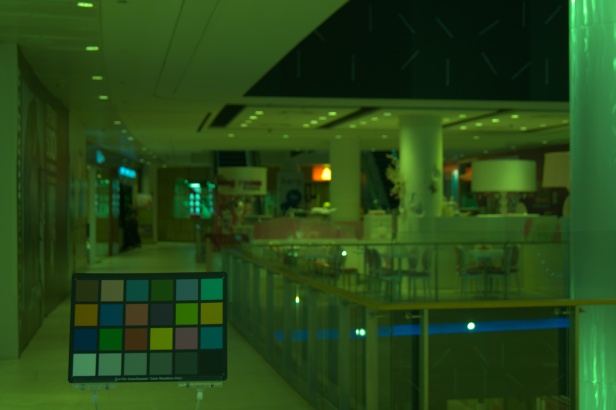}
}
\subfloat[(b) FFCC (error: $12.27\degree$)]
{
  \centering
  \includegraphics[width=0.03\columnwidth,height=2.55cm]{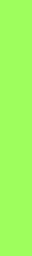}
  \hspace*{-1.5mm}
  \includegraphics[width=0.22\textwidth]{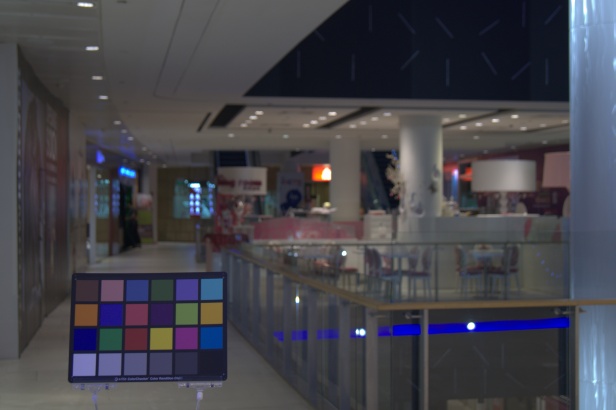}
}
\subfloat[(c) Ours (error: $10.23\degree$)]
{
  \centering
  \includegraphics[width=0.03\columnwidth,height=2.55cm]{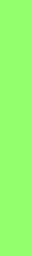}
  \hspace*{-1.5mm}
  \includegraphics[width=0.22\textwidth]{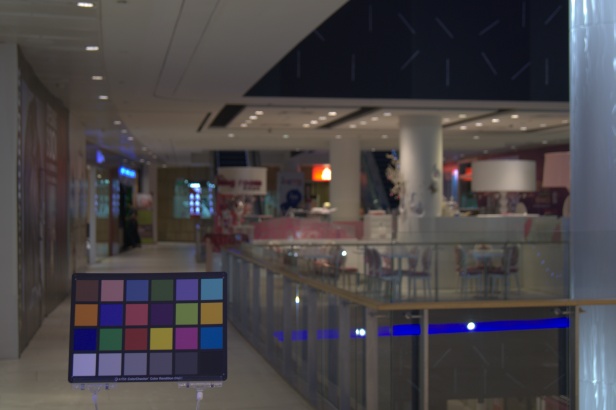}
}
\subfloat[(d) Ground Truth]
{
  \centering
  \includegraphics[width=0.03\columnwidth,height=2.55cm]{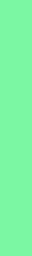}
  \hspace*{-1.5mm}
  \includegraphics[width=0.22\textwidth]{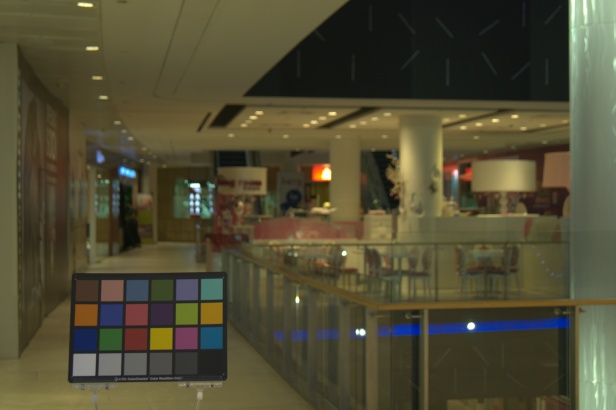}
}

\subfloat[(a) Input image]
{
  \centering
  \includegraphics[width=0.22\textwidth]{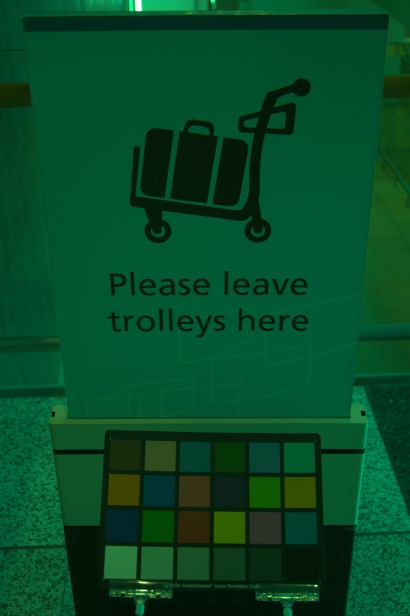}
}
\subfloat[(b) FFCC (error: $9.36\degree$)]
{
  \centering
  \includegraphics[width=0.03\columnwidth,height=5.75cm]{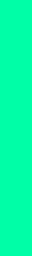}
  \hspace*{-1.5mm}
  \includegraphics[width=0.22\textwidth]{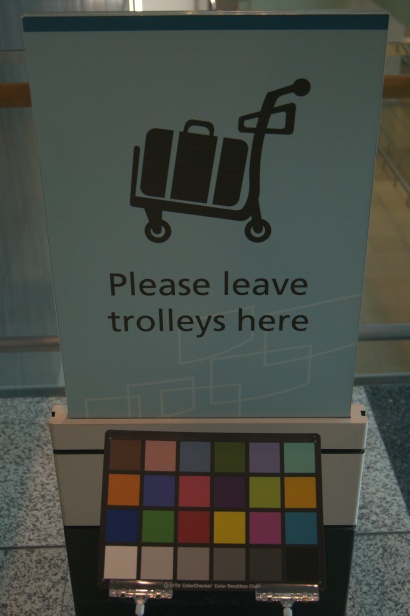}
}
\subfloat[(c) Ours (error: $11.21\degree$)]
{
  \centering
  \includegraphics[width=0.03\columnwidth,height=5.75cm]{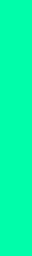}
  \hspace*{-1.5mm}
  \includegraphics[width=0.22\textwidth]{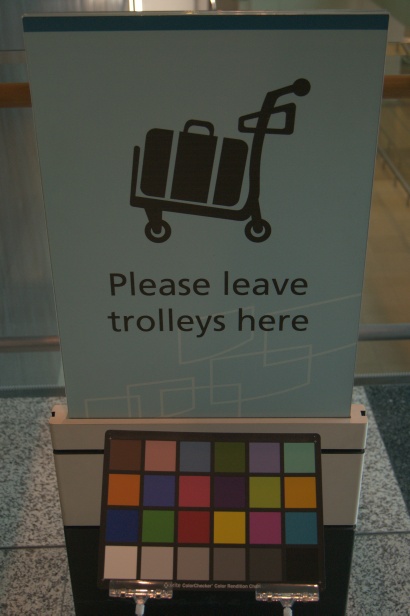}
}
\subfloat[(d) Ground Truth]
{
  \centering
  \includegraphics[width=0.03\columnwidth,height=5.75cm]{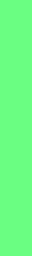}
  \hspace*{-1.5mm}
  \includegraphics[width=0.22\textwidth]{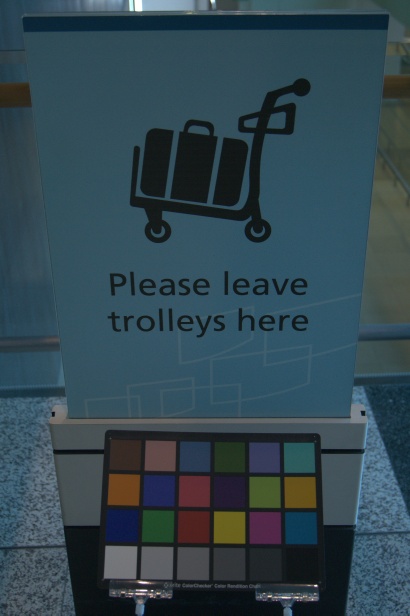}
}

\subfloat[(a) Input image]
{
  \centering
  \includegraphics[width=0.22\textwidth]{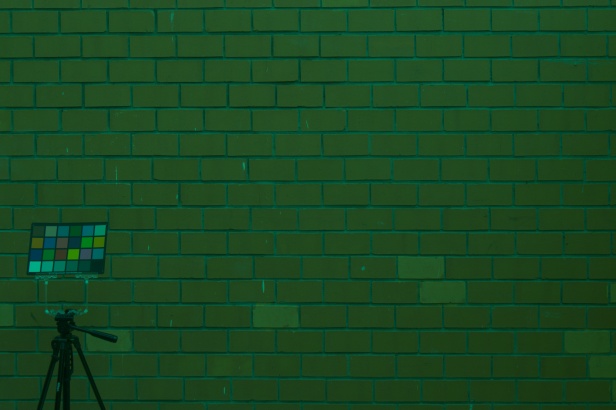}
}
\subfloat[(b) FFCC (error: $10.37\degree$)]
{
  \centering
  \includegraphics[width=0.03\columnwidth,height=2.55cm]{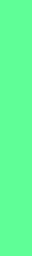}
  \hspace*{-1.5mm}
  \includegraphics[width=0.22\textwidth]{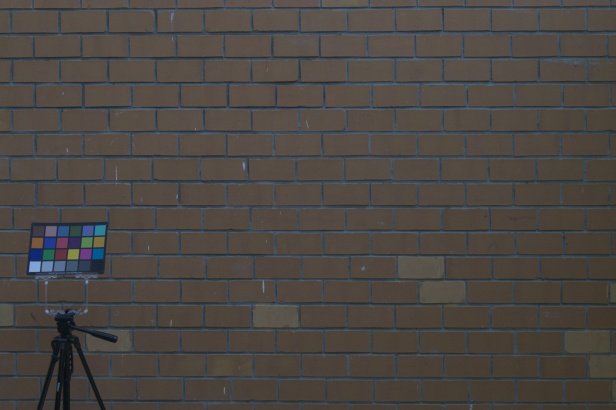}
}
\subfloat[(c) Ours (error: $7.49\degree$)]
{
  \centering
  \includegraphics[width=0.03\columnwidth,height=2.55cm]{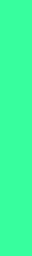}
  \hspace*{-1.5mm}
  \includegraphics[width=0.22\textwidth]{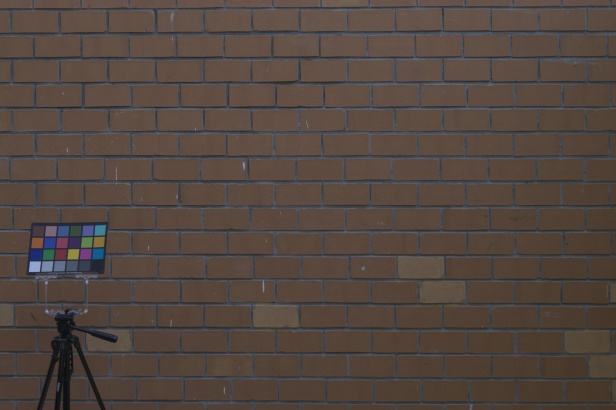}
}
\subfloat[(d) Ground Truth]
{
  \centering
  \includegraphics[width=0.03\columnwidth,height=2.55cm]{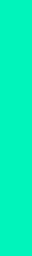}
  \hspace*{-1.5mm}
  \includegraphics[width=0.22\textwidth]{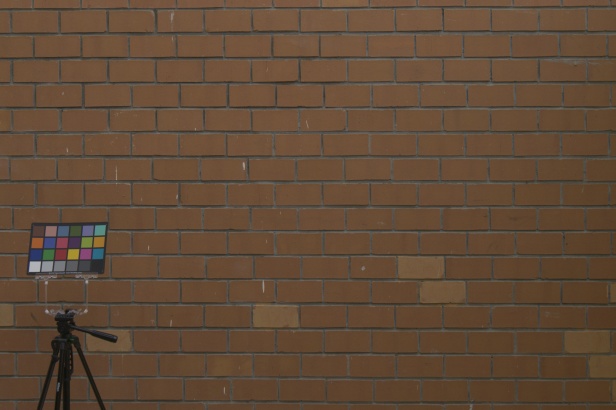}
}
\caption{\acceptedrev{Visual comparison of FFCC~\cite{barron2017fast} and our method with Sony dataset (NUS~\cite{cheng2014illuminant}). We select the five \textbf{largest} combined mean angular error to explore method behaviour for images that are commonly challenging. Images are rendered in sRGB color space.}}
\label{fig:worst_results}
\end{figure*}

We provide further extremely challenging examples in~\cref{fig:worst_results}. We explicitly select the five largest combined mean angular-error images. We observe that our method shows consistently strong performance and also highlight that these samples constitute cases of both ambiguous and multi-illuminant scenes, breaking the fundamental global illuminant assumption (made by both methods).



\end{document}